\documentclass[lettersize,journal]{IEEEtran}
\usepackage{amsmath,amsfonts}
\usepackage{array}
\usepackage[caption=false,font=normalsize,labelfont=sf,textfont=sf]{subfig}
\usepackage{textcomp}
\usepackage{stfloats}
\usepackage{url}
\usepackage{verbatim}
\usepackage{graphicx}
\usepackage{cite}

\usepackage{makecell,multirow}
\usepackage{threeparttable}
\usepackage{amsmath}
\usepackage{amssymb}
\usepackage{latexsym}
\usepackage{CJK}
\usepackage{booktabs}
\usepackage{tabularx}

\usepackage{bigstrut}
\usepackage{bm}
\usepackage{epstopdf}

\usepackage[lined,boxed,commentsnumbered,ruled]{algorithm2e}
\newsavebox\CBox

\usepackage[pagebackref=true,breaklinks=true,colorlinks,bookmarks=false]{hyperref}

\begin{document}
\title{Rethinking Iterative Stereo Matching from Diffusion Bridge Model Perspective}
\author{Yuguang Shi
\thanks{Yuguang Shi is with the School of Automation, Southeast University and the Key Laboratory of Measurement and Control of Complex Systems of Engineering, Ministry of Education, Nanjing 210096, China(e-mail: syg@seu.edu.cn).
	
}

}
\markboth{Journal of \LaTeX\ Class Files,~Vol.~14, No.~8, August~2021}%
{Shell \MakeLowercase{\textit{et al.}}: A Sample Article Using IEEEtran.cls for IEEE Journals}

\IEEEpubid{\begin{minipage}{\textwidth}\ \\[30pt] \centering
		Copyright \copyright 20xx IEEE. Personal use of this material is permitted. 
		However, permission to use this material for any other purposes must \\ be obtained 
		from the IEEE by sending an email to pubs-permissions@ieee.org.
\end{minipage}}

\maketitle

\begin{abstract}
	Recently, iteration-based stereo matching has shown great potential. However, these models optimize the disparity map using RNN variants. The discrete optimization process poses a challenge of information loss, which restricts the level of detail that can be expressed in the generated disparity map. In order to address these issues, we propose a novel training approach that incorporates diffusion models into the iterative optimization process. We designed a Time-based Gated Recurrent Unit (T-GRU) to correlate temporal and disparity outputs. Unlike standard recurrent units, we employ Agent Attention to generate more expressive features. We also designed an attention-based context network to capture a large amount of contextual information. Experiments on several public benchmarks show that we have achieved competitive stereo matching performance. Our model ranks first in the Scene Flow dataset, achieving over a $7\%$ improvement compared to competing methods, and requires only 8 iterations to achieve state-of-the-art results.
\end{abstract}

\begin{IEEEkeywords}
Stereo Matching, attention, diffusion models.
\end{IEEEkeywords}

\section{Introduction}
\IEEEPARstart{D}{epth} estimation from stereo images is a long-standing task in the field of computer vision. The main objective of the task is to estimate a pixel-wise displacement map, also known as disparity, which can be utilized to determine the depth of the pixels in the scene \cite{zhao2023high}. It is a key component in a variety of downstream issues, showing significant success in autonomous driving, augmented reality, and novel view generation.

\begin{figure}[!t]
	\centering
	\includegraphics[width=3.5in]{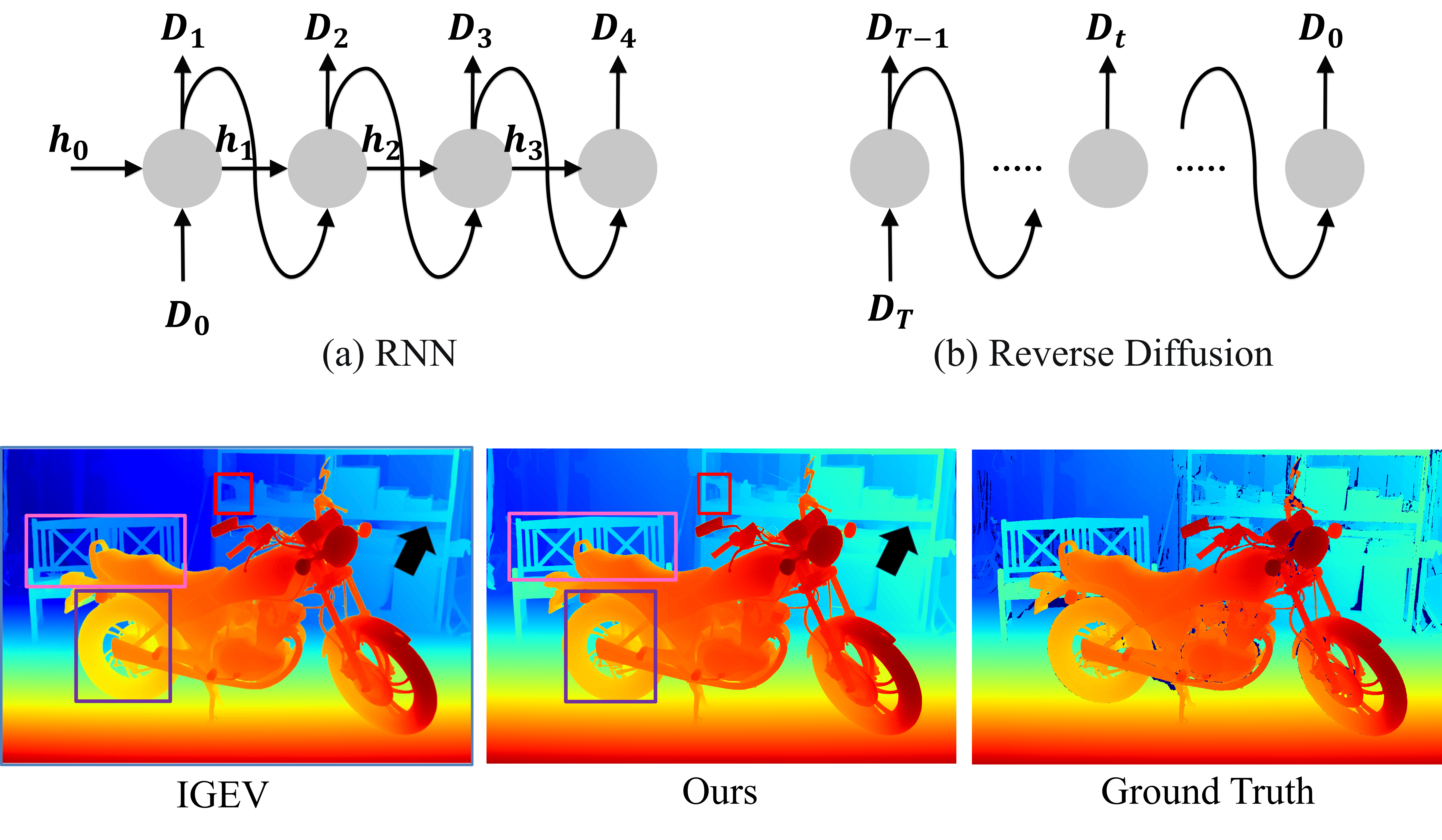}
	\caption{Row 1: Comparison of RNN's recursive multi-step prediction and the diffusion models reverse process. Row 2: Visual comparison with IGEV on middlebury dataset.}
	\label{fig_1}
\end{figure}
 
\begin{figure*}[t]
	\centering  
	\includegraphics[width=18cm,height=6.5454cm]{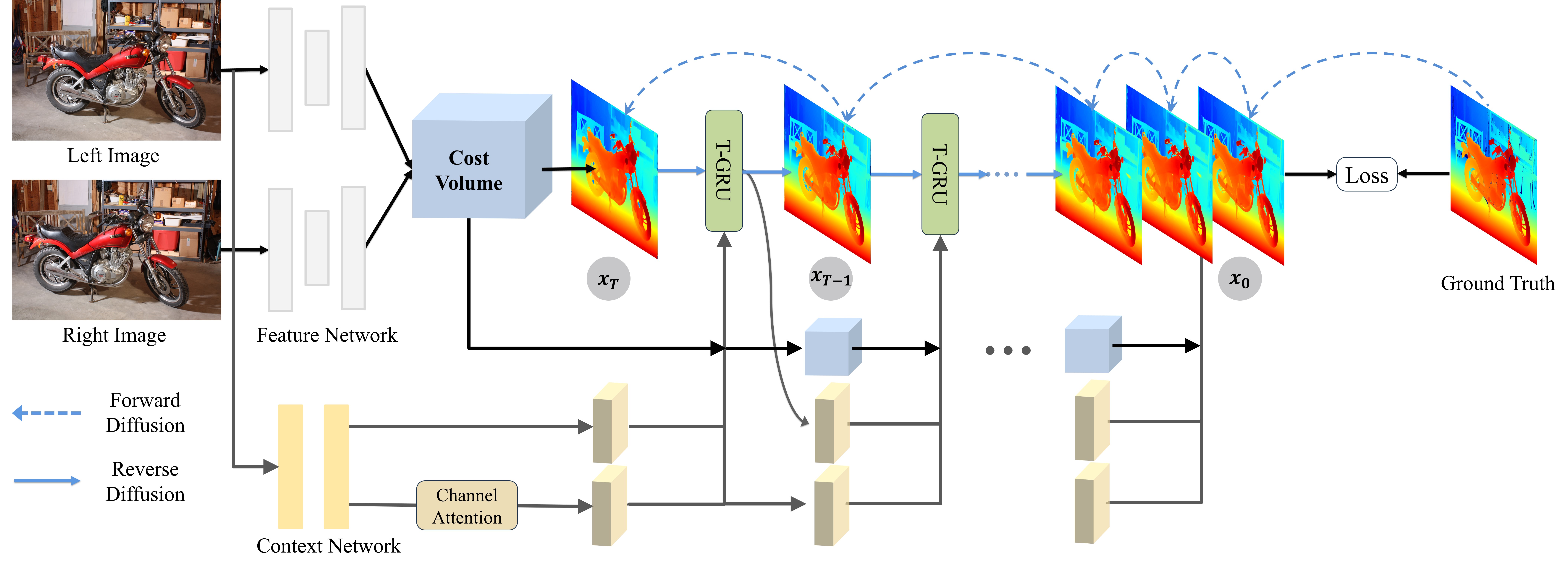}
	\caption{Overview of our proposed DMIO. DMIO main contribution consists of three main modules, which are attention-based context network, bridge diffusion disparity refinement, and T-GRU based update operator.}
	\label{fig_2}
\end{figure*}
Due to homogeneity in inherently ill-posed regions, such as occlusion areas, repeated patterns, textureless regions, and reflective surfaces, accurately matching corresponding points is quite challenging. To address this issue, conventional stereo methods commonly rely on regularization of cost volume and regularization techniques. This paradigm is broken by iterative optimization-based methods, which extract information from a high-resolution cost volume and utilize variations of Recurrent Neural Networks (RNNs) like Long Short-Term Memory (LSTM) and Gated Recurrent Unit (GRU) to iteratively enhance the disparity map. From the perspective of the entire iterative optimization sequence, the characteristic of RNNs is that their complexity grows linearly with the increase in spatial resolution. Therefore, iteration-based methods facilitate faster inference speeds while bypassing complex cost aggregation operations.

However, there are some limitations to iterative methods. As shown in the first row (a) of Fig. 1, an RNN is a natural stateful model for sequential data, which makes it impossible to train in parallel and increases training time. Furthermore, the number of recurrent steps is limited to a maximum number to reduce computational costs. Consequently, the intermediate disparity map between training and generation becomes discrete, potentially resulting in gaps in local edges and details.

Recently, diffusion models have gained sudden prominence. The first row (b) of Figure 1 illustrates the sampling process of a diffusion model. It begins with random noise and progresses iteratively by using a neural network to generate predictions and updating the image according to these predictions. The process resembles an RNN. The difference is that the RNN has a hidden state that is repeatedly passed through the GRU to obtain updates, whereas the GRU allows to be viewed as a specialized network in the diffusion model. In terms of training, the advantage of the diffusion model is to avoid backpropagation through recurrence at all (Sect. 3.1), so in a sense, this parallel approach provides a more scalable training process that compensates for the shortcomings of RNNs.

In this paper, we argue to provide a rethinking of the iterative optimization process of stereo matching methods from the perspective of diffusion models. To be specific, we draw inspiration from recent works \cite{li2023bbdm,liu2022flow} that directly model the transport between two arbitrary probability distributions. We propose a stereo matching method called Diffusion Models for Iterative Optimization (DMIO).

DMIO is a novel stereo matching training method, and the network architecture is shown in Figure 2. First, the weight-sharing backbone network extracts consistent features from both left and right images. The stereo features are structured as a 4D cost volume, which is fed into a 3D CNN for normalization and disparity regression, resulting in an initial estimate of the disparity. Second, following \cite{lipson2021raft}, we introduce a context network extraction feature in the left image, which is utilized for iterative optimization. The difference is that we add an additional channel attention feature extractor in the second branch, which is designed to capture long-range pixel dependencies and preserve high-frequency information. Finally, we introduce a bridge diffusion process without random noise prior distribution in the iterative optimization module. The limitation of the standard diffusion approach is that the U-Net is too heavy to be suitable for a lightweight iterative module. Therefore, we propose a Time-based Gated Recurrent Unit (T-GRU), which includes a time encoder and an optional agent attention mechanism.






Our main contributions are summarized as follows:
\begin{itemize}
	\item[$\bullet$]This work proposes a novel stereo matching training method that reformulates iterative optimization as an image-to-image translation (I2IT) diffusion model. This provides a new direction for the application of diffusion models.
	\item[$\bullet$]We propose a novel iterative update operator T-GRU for iterative refinement-based stereo matching methods.	
	\item[$\bullet$]Detailed experiments on SceneFlow, KITTI 2012, KITTI 2015, ETH3D and Middlebury datasets show that our method performs very competitively.
\end{itemize}

\section{RELATED WORK}

\noindent{\bf Cost Volume-based Stereo Matching Methods:} Matching cost aggregation arguably is one of the core steps in stereo matching algorithm. The recent cost volume-based methods inherit the ideas from traditional matching most computation, rely on 3D convolutional networks instead of handcrafted schemes. Benefiting from large-scale training data and end-to-end training \cite{zhang2020adaptive}, deep learning-based stereo methods have achieved the state-of-the-art results, which has become a convention of modern stereo matching such as \cite{kendall2017end,chang2018pyramid,duggal2019deeppruner,cheng2020hierarchical,wang2021pvstereo,shen2022pcw}.
GCNet\cite{kendall2017end} is a firstly proposed 3D encoder-decoder architecture, which regressing disparity from a cost volume built by a pair of stereo features. PSMNet\cite{chang2018pyramid} further proposes a stacked hourglass 3D CNN and pyramid pooling module refine context information in concatenation volume. GwcNet\cite{guo2019group} adopts a propose group-wise correlation to construct cost volumes. DeepPruner\cite{duggal2019deeppruner} propose a algorithm to obtain a sparse representation of the cost volume based on the coherent nature of the world. LEAStereo\cite{cheng2020hierarchical} utilize the NAS technique to select the optimal structures for 3D cost volume. GANet\cite{zhang2019ga} designs a semi-global aggregation layer and a local guided aggregation layer to further improve the accuracy. ACVNet\cite{xu2022attention} propose attention concatenation volume to produce more accurate similarity measures. Another family methods \cite{cheng2020deep,yang2020cost,gu2020cascade,mao2021uasnet} are based on cascade pyramid volume and employ each volume to estimate high-quality disparity maps separately. PCWNet\cite{shen2022pcw} follows the same architecture, which fusing multi-scale combination volumes to extract domain-invariant features.

However, the large 3D convolutional neural networks (3D CNNs) are often limited by high computing quantities, so it is necessary to use about one second to infer a disparity map from a pair of rectified stereo images. One popular real-time approach is to reduce the computation of the cost volume by downsampling methods\cite{khamis2018stereonet,zhang2018activestereonet} that pipline first downsamples the cost volume and then relies on a carefully designed upsampling layers\cite{xu2021bilateral} to recover fine details. In particular, HITNet\cite{tankovich2021hitnet} integrating image warping, spatial propagation and high-resolution initialization step into the network architecture, and performs competitively with sophisticated methods. Recent method ACVNet\cite{xu2023accurate} proposes a concise cost volume representation and does not require additional propagation loss, skew loss and confidence loss for training.

\noindent{\bf Transformer-based Stereo Matching Methods:} With the support of attention mechanisms, a majority of state-of-the-art deep stereo matching methods are based on transformer. STTR\cite{li2021revisiting} takes the first attempt to use alternating self-cross attention modules to estimate the disparity and corresponding occlusion mask from an aspect of the transformer. CSTR\cite{guo2022context} employs a plugin module, Context Enhanced Path, to help better integrate global information and deal with hazardous regions, such as texturelessness, specularity, or transparency. In order to alleviate the problem that the transformer-based methods perform not perform well in regions with local texture details, ELFNet\cite{lou2023elfnet} takes the approach of fusing transformer and cost volume to capture complementary information and thus improve the performance.

\begin{figure}[t]
	\centering
	\includegraphics[width=8cm,height=5.7cm]{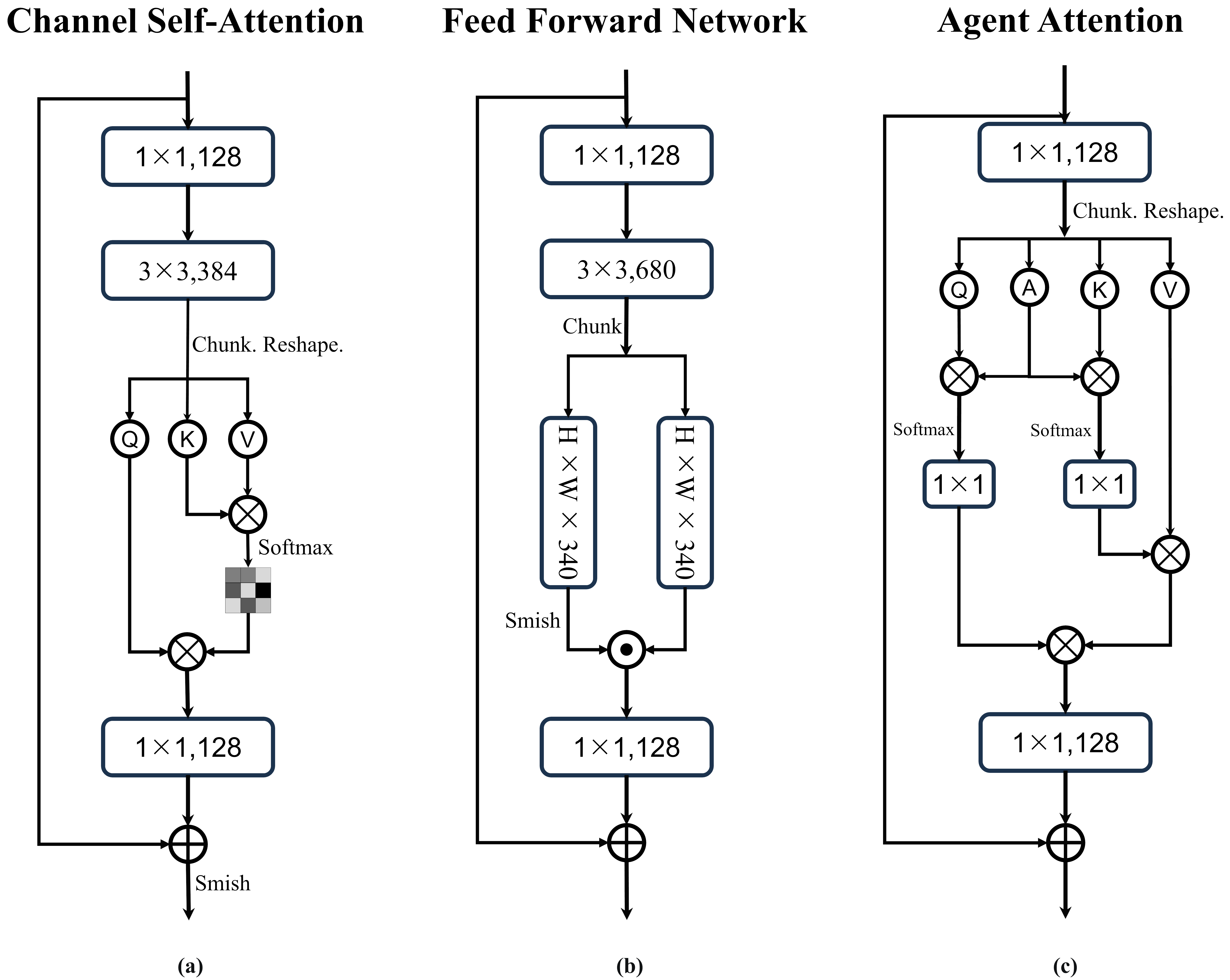}
	\caption{Block designs for a Channel Self-Attention, a Feed-Forward Network, and a Agent Attention.}	
	\label{fig_6}
\end{figure}

\noindent{\bf Iterative Refinement-based Stereo Matching Methods:} 
In recent years, deep learning-based iterative refinement has made significant progress in optical flow. The most popular iterative optimization-based method might be RAFT \cite{teed2020raft}, which constructs a 4D multi-scale correlation volume and utilizes a convolutional GRU block as an update operator. Similar to RAFT, RAFT-Stereo \cite{lipson2021raft} using adopts multi-level ConvGRUs to update disparity map iteratively, which is the first application of iterative methods in the field of stereo matching. IGEV-Stereo \cite{xu2023iterative} provides a better initial disparity map than RAFT-Stereo \cite{lipson2021raft} by introducing a Geometry Encoding Volume. DLNR\cite{zhao2023high} proposes to replace the traditional GRU with an Decouple LSTM and normalize it to maintain fine details. Li et al. \cite{li2022practical} introduce adaptive group correlation layer into the iteration module, and use cross attention mechanism to reduce the matching ambiguity in stereo image pairs. GOAT \cite{liu2024global} propose to replace the cost volume with the vision transformer and attention mechanism to calculate the initial disparity, which improves the accuracy of the occluded region. CREStereo++ \cite{jing2023uncertainty} introduce variance-based uncertainty estimation and deformable convolution network in iterations to enhance the robustness in different scenarios. DynamicStereo \cite{karaev2023dynamicstereo} propose 3D-GRU to estimate disparity for stereo videos



\noindent{\bf Diffusion Model:} 
At present, diffusion models have achieved great success in image generation, and thus exploring its potential in downstream tasks has become one of the most widely concerned studies, which include detection\cite{chen2023diffusiondet}, segmentation\cite{amit2021segdiff,baranchuk2021label}, image to image translation\cite{choi2021ilvr,kawar2022denoising,li2023bbdm}, super resolution\cite{saharia2022image,lin2023diffbir}, image editing\cite{meng2021sdedit}, video generation\cite{ho2022video}, etc. There has even been some progress at applying diffusion models to monocular depth estimation\cite{duan2023diffusiondepth,saxena2023monocular,ke2023repurposing,saxena2023zero,shao2023monodiffusion,saxena2024surprising}. To this end, We are motivated to further explore the potential for application in stereo matching. 




\begin{figure}[!t]
	\centering
	\includegraphics[width=3.5in]{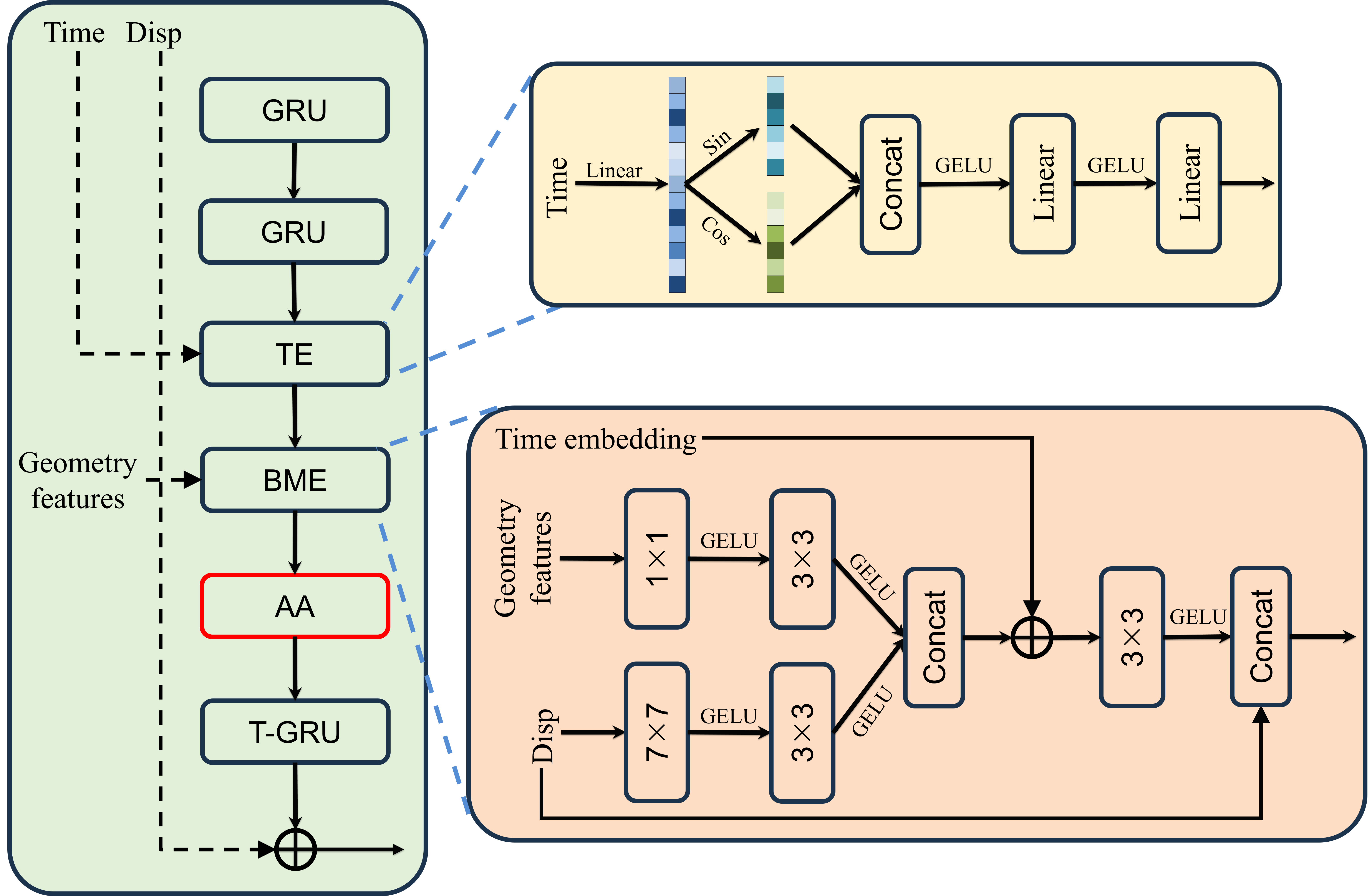}
	\caption{The diffusion network architecture consists of the Gated Recurrent Unit (GRU), Time Encoder (TE), Basic Motion Encoder (BME), optional Agent Attention (AA), and Time-based Gated Recurrent Unit (T-GRU).}
	\label{fig_4}
\end{figure}

\section{THE PROPOSED METHOD}
Figure 2 shows the architecture of DMIO. It inherits the influence flow estimation pipeline of IGEV, which consists of a weight-sharing feature network, a cost volume, an attention-based context network, a bridge diffusion disparity refinement, a T-GRU-based update operator, and a spatial upsampling module.

\subsection{Feature Extractor and Cost Volume}
\noindent{\bf Feature Network.} For the trade-off between speed and accuracy, and for a fair comparison with other methods, we adopt the same MobileNetV2 network as IGEV \cite{xu2023iterative} to extract feature maps. Details about the multi-scale features used to construct the cost volume can be referred to in \cite{xu2023iterative}.

\noindent{\bf Cost Volume.}
Our cost volume module is derived from IGEV \cite{xu2023iterative} with combining the $\mathbf{C} _ { G } $ and $\mathbf{C} _ {A } $ two different pyramids. Given the left and right features $(f_{l1} ,f_{r1} )$ extracted from the stereo image, a group-wise correlation volume $\mathbf{C} _ { G } $ is first constructed :
\begin{eqnarray}
\mathbf{C}_ { c o r r } ( g , d , x , y ) = \frac { 1 } { N _ { c } / N _ { g } } \langle f_{l1 } ^ { g } ( x , y ) , f _ {r1} ^ { g } (x - d , y) \rangle
\end{eqnarray}
Group-wise correlation volume works by computes the group of correlation maps after splits features. Where $\left \langle \cdot ,\cdot  \right \rangle $ is the inner product, $d$ is all disparity levels,
$g$ is feature groups, $N _ { c }$ denotes the channels of unary features, $N _ { g }$ is the number of groups, and each feature group therefore has $N_{ c } / N_{g}$ channels. The geometry encoding volume $\mathbf{C} _ { G } $ is obtained from $\mathbf{C}_ { c o r r }$, which is further processed by a lightweight 3D regularization network $\mathbf{R}$, for details on $\mathbf{R}$, see \cite{guo2019group} and \cite{xu2023iterative}, respectively : 
\begin{eqnarray}
\mathbf{C} _ { G } = \mathbf{R} (\mathbf{ C} _ { c o r r } )
\end{eqnarray}
Initial starting disparity $\mathbf{d} _ { 0} $ is computed from the $\mathbf{C} _ { G } $ :
\begin{eqnarray}
\mathbf{d} _ { 0 } = \sum _ { d = 0 } ^ { D - 1 } d \times Softmax (\mathbf{C} _ { G }( d ) )
\end{eqnarray}
where $d$ is a predetermined set of disparity indices at $1/4$ resolution.

Second, given the left and right features $(f_{l2} ,f_{r2} )$, an all-pair correlation cost volume $\mathbf{C} _ {A }$ is constructed:
\begin{eqnarray}
\mathbf{C}_{A}=\sum_{h} \mathbf{f}_{l2} \cdot \mathbf{f}_{r2}
\end{eqnarray}

\subsection{Attention-based Context Network}
Following RAFT-Stereo \cite{lipson2021raft}, when provided with a left image map, the context encoder comprises a sequence of ResNet-like blocks \cite{resnet} and downsampling layers to extract features at $1/4$ resolution. Additionally, three extra downsampling layers are utilized to acquire multi-level context features at $1/4$, $1/8$, and $1/16$ resolution, each with $128$ channels. Initial hidden and context information is obtained by the $Tanh$ and $ReLU$ activation functions, respectively.

Inspired by the recent success of GMA \cite{jiang2021learning} and DLNR \cite{zhao2023high}, we introduce a channel self-attention and feed-forward network (refer to Fig. 3(a) and 3(b)) and present the impact of these designs on quality in ablation experiments. The key ingredient of channel self-attention is to compute cross-covariance across channels to generate an attention map that encodes the global context implicitly. Given a feature map $X$ of context information, channel self-attention further aggregates pixel-wise cross-channel context by applying a $1\times1$ convolution.  Then a $3\times3$ depth-wise convolution is applied to generate query (Q), key (K), and value (V) projections. The query and key matrix dot products interact to generate a transposed attention map, denoted as A. Then output feature map $\hat{\mathbf{X}}$ can be expressed as follows:
\begin{eqnarray}
\begin{array}{l}
\hat{\mathbf{X}}=W_{p} \operatorname{Attention}(\hat{\mathbf{Q}}, \hat{\mathbf{K}}, \hat{\mathbf{V}})+\mathbf{X}, \\
\operatorname{Attention}(\hat{\mathbf{Q}}, \hat{\mathbf{K}}, \hat{\mathbf{V}})=\hat{\mathbf{V}} \cdot \operatorname{Softmax}(\hat{\mathbf{K}} \cdot \hat{\mathbf{Q}} / \alpha)
\end{array}
\end{eqnarray}
where $W_{p}$ is a $1\times1$ point-wise convolution, $\alpha$ is a learnable scaling parameter used to control the magnitude of the dot product of $K$ and $Q$ before applying the softmax function. 

The final output is activated with the SMISH \cite{wang2022smish} non-linearity:
\begin{eqnarray}
\hat{\mathbf{X}} =SMISH(\hat{\mathbf{X}}) = \hat{\mathbf{X}} \cdot \tanh \left[ \log ( 1 + s i g m o i d (\hat{\mathbf{X}}) ) \right] 
\end{eqnarray}
The SMISH is capable of reducing the inefficiency in training optimization. To transform features, we take an additional step after channel self-attention by utilizing a feed-forward network that incorporates the SMISH activation function. Please refer to Restromer \cite{zamir2022restormer} for more details.

\subsection{Bridge Diffusion Disparity Optimization}
The forward diffusion process of T-step Denoising Diffusion Probabilistic Models (DDPM) \cite{ho2020denoising} begins with clean data and ends with gaussian noise \cite{li2023bbdm}. However, DDPM is not well-suited for image transformation tasks across two distinct domains. Following Rectified Flow \cite{liu2022flow} and DDBM \cite{zhou2023denoising}, our goal in Disparity Optimization is to establish a bridge diffusion model that differs from DDPM. To avoid errors caused by noise, this model does not introduce gaussian noise but directly establishes a mapping between the initial disparity domain obtained in the cost volume and the ground truth disparity domain.

\noindent{\bf Forward Process.} Given initial disparity $\mathbf{D} _ {0} \in \mathbb{R}^{\frac{W}{4} \times \frac{H}{4} \times 1}$ and destination disparity $\mathbf{D}_ {gt} \in \mathbb{R}^{ W \times H \times 1}$. Our method formulates the sampling trajectory process as an ordinary differential equation (ODE):
\begin{eqnarray}
\underbrace{\mathrm{d} \mathbf{D}_{t}}_{\text {drift }}=\underbrace{v_{\theta}\left(\mathbf{D}_{t}, t\right)}_{\text {velocity }} \underbrace{\mathrm{d} t}_{\text {time interval }} \text {, with } t \in[0,1] \text {. }
\end{eqnarray}
where $\mathbf{D}_{t}$ is the intermediate disparity states at time $t$ and the velocity filed $ v _ { \theta } : \mathbb{R}  \rightarrow \mathbb{R} $ is a neural network with $\theta $ as its parameters, learned by minimizing a simple mean square objective:
\begin{eqnarray}
\begin{array}{l}
\min _{\theta} \int_{0}^{1} \mathbb{E}\left[\|\left(v_{\theta}\left(\mathbf{D}_{t}, t\right)-\left(\mathbf{D}_{gt}-\mathbf{D}_{0}\right) \|^{2}\right] \mathrm{d} t \right. 
\end{array}
\end{eqnarray}
where $\mathbf{D}_{t}$  is a linear interpolation of $\mathbf{D}_{0}$ and $\mathbf{D}_{gt}$, it can be summarized as follows:

\begin{eqnarray}
\begin{aligned}
\boldsymbol{\mathbf{D}}_{t} & =\sqrt{\left(1-\beta_{t}\right)} \boldsymbol{\mathbf{D}}_{gt}+\sqrt{\beta_{t}} \boldsymbol{\mathbf{D}_ {0}} \\
\boldsymbol{\mathbf{D}}_{t-1} & =\sqrt{\left(1-\beta_{t-1}\right)} \boldsymbol{\mathbf{D}}_{gt}+\sqrt{\beta_{t-1}} \boldsymbol{\mathbf{D}_ {0}}
\end{aligned}
\end{eqnarray}
$\mathbf{D}_{t}$ is determined by various noisy scheduling algorithms, where $\beta$ represents the hyper-parameter sequence determined by the algorithm.

\noindent{\bf Reverse Process.} In the reverse process of our method, the diffusion process begins with the initial disparity map and progressively removes inaccurate disparity estimates to achieve the final disparity map. When updating the disparity map, discretizing the ODE in Equation 1 needs to be approximated with a numerical solver, such as the forward Euler solver, denoted as:
\begin{eqnarray}
\mathbf{D}_{(\hat{t}+1) / N}^{\prime} \longleftarrow \mathbf{D}_{\hat{t} / N}^{\prime}+\frac{1}{N} v_{\theta}\left(\mathbf{D}_{\hat{t} / N}^{\prime}, \frac{\hat{t}}{N}\right),
\end{eqnarray}
where $\mathbf{D}_ {t} ^ { \prime } \in \mathbb{R}^{\frac{W}{4} \times \frac{H}{4} \times 1}$ denotes our generated disparity maps and $ \mathbf{D}_ { 0 } ^ { \prime } =\mathbf{D}_ { 0 }$, $N$ denotes numbers of sampling step. The integer time step $\hat{t}$ is defined as $ \hat { t } \in \left\{ 0 , 1 , \cdots , N - 1 \right\}$. In practice, due to the unique characteristics of the GRU model and for better alignment with current iteration-based methods, the disparity update is as follows:
\begin{eqnarray}
\mathbf{D}_{(\hat{t}+1) / N}^{\prime} \longleftarrow \mathbf{D}_{0}^{\prime}+ v_{\theta}\left(\mathbf{D}_{\hat{t} / N}^{\prime}, \frac{\hat{t}}{N}\right),
\end{eqnarray}
Obviously, choosing parameters $N$ results in a trade-off between cost and accuracy. A larger $N$ provides a better approximation of the ODE but also leads to higher computational costs \cite{liu2023instaflow}. After the reverse process, the updated 1/4 resolution disparity map is upsampled to full resolution to obtain the final disparity map.

\begin{algorithm}
	\label{alg1}	
	\caption{Sigmoid Noise schedule.}
	\LinesNumbered 
	\KwIn{Time $t$; start=-3; end=3; $\tau$ =1; MIN=1e-9;}
	v\_start  = sigmod(start / $\tau$)\\
	v\_end  = sigmod(end / $\tau$)\\
	output = sigmoid((t $\times$ (end - start) + start) / $\tau$)\\
	output = (v\_end - output) / (v\_end - v\_start)\\
	\Return{$\beta$ = np.clip(output, MIN, 1.)}	
\end{algorithm}

\noindent{\bf Noise Schedule.} Linear schedule \cite{ho2020denoising} and cosine schedule \cite{nichol2021improved} are two classic and widely used noise schedule strategies. The latest research by Google \cite{chen2023importance} pointed out that noise scheduling is vital to performance. In order to improve the stability of the training model, as shown in Algorithm 1, we have opted for a more robust sigmoid schedule strategy \cite{j2022scalable}. By adjusting the hyperparameters of sigmoid functions, it leads to different noise schedules.

\subsection{T-GRU-based Update Operator} As mentioned above (Section 3.3), DMIO uses a velocity neural network $v_{\theta}$ to predict the distribution $\mathbf{D}_{t-1}^{\prime}$. The role of the network in diffusion models is crucial. A popular example is the U-Net \cite{ronneberger2015u}, which is based on CNN \cite{ho2020denoising,si2023freeu,williams2024unified} or Transformer \cite{bao2023all,peebles2023scalable} blocks, and has a high computational complexity. Since feature extractor and cost volume already account for the majority of the computation in DMIO, optimization tasks typically require less inference time. Following DiffusionDepth's proposal \cite{duan2023diffusiondepth}, our idea is to enable the diffusion network to deviate from the U-Net structure design towards a more lightweight configuration. 

As shown in Figure 4, we introduce a T-GRU-based update operator to implement this network. Following $ \mathbf{D}_{0}^{\prime}$, the network has a stacked structure, comprising two GRUs, a time encoder, a basic motion encoder, an optional agent attention, and a T-GRU. In the same structure as RAFT-Stereo \cite{lipson2021raft}, the top-to-bottom three levels of ConvGRUs take as input the hidden states and context features at resolutions of 1/16, 1/8 and 1/4, respectively. At 1/4 resolution, T-GRU specifically incorporates the disparity and BME output as additional inputs. After T-GRU, we use two additional convolutional heads to decode the disparity features at each time step. 

\noindent{\bf Time Encoder.} DMIO must incorporate relative or absolute positional information related to time markers into the model to achieve sequential diffusion. Therefore, we add a TE to the network stack. During training, the input $t \in[0,1]$ is continuous, whereas during testing, it is discrete. As shown in Figure 4, the initial layer is a basic linear embedding. Then, we utilize the transformer's sinusoidal position embedding \cite{vaswani2017attention}, dividing the time encoding into two parts by employing sine and cosine functions with different frequencies. This approach enables the model to effectively learn relative positions. After concatenation, we add two GELU activation function \cite{hendrycks2016gaussian} and linear layers to adjust the sequence dimension.

\noindent{\bf Basic Motion Encoder.} Since the disparity is updated at each iteration in the reverse process. Similarly to IGEV \cite{xu2023iterative}, we use the current disparity $\mathbf{D}_{t}^{\prime}$ to index the combined geometry encoding volume via linear interpolation, resulting in a set of current step geometric features $\mathbf{C} _ {F} $, which can be expressed as follows:
\begin{eqnarray}
\mathbf{G}_{F}= \sum _ { i = - r } ^ { r } \operatorname{Concat}[\mathbf{C}_{G}(\mathbf{D}_{t}^{\prime}+ i ) ,\mathbf{C}_{A} (\mathbf{D}_{t}^{\prime}+ i ) , \\ \mathbf{C}_{G}^ {p} ( \mathbf{D}_{t}^{\prime} / 2 + i ) ,  \mathbf{C}_{A}^ { p } (\mathbf{D}_{t}^{\prime} / 2 + i ) ] ,
\end{eqnarray} 
where $r$ is the indexing radius, $p$ is the pooling operation. As shown in Figure 4, the basic motion encoder takes current step geometric features $\mathbf{C} _ {F} $ and disparity prediction $\mathbf{D}_{t}^{\prime}$ as input. It obtains the output encoding $x_{k}$ through two convolutional layers, which then successively process additional conditional information, time embedding $\mathbf{t}_{e}$, and $\mathbf{D}_{t}^{\prime}$.
\begin{eqnarray}
x_{k} =\left[\operatorname{Encoder}_{g}\left(\mathbf{G}_{F}\right)+\mathbf{t}_{e}, \operatorname{Encoder}_{d}\left(\mathbf{d}_{k}\right)+\mathbf{t}_{e}, \mathbf{d}_{k} \right] 
\end{eqnarray}

\begin{figure}[!t]
	\centering
	\includegraphics[width=3.5in]{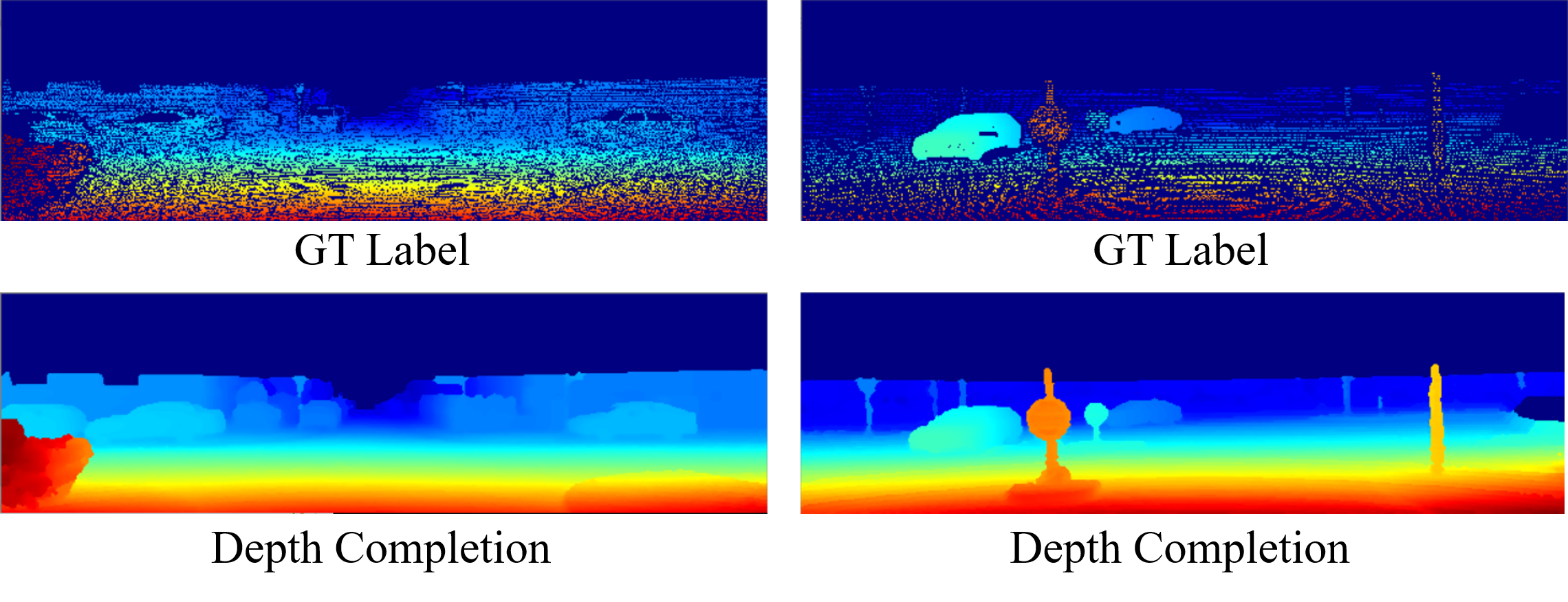}
	\caption{Infilling ground truth disparity missing values using interpolation.}
	\label{fig_1}
\end{figure}

\noindent{\bf Agent Attention.} In order to enhance the ability to represent features and reduce ambiguity caused by occlusion, we incorporate agent attention after BME. Agent attention combines Softmax and linear attention, offering the advantages of linear complexity and high expressiveness. The detailed graph of agent attention is shown in Figure 3(c), we acquire newly defined agent tokens $A$ through pooling. Subsequently, the agent token is used to aggregate information from $V$, and $Q$ queries features from the agent feature. It is defined as follows:

\begin{eqnarray} 
\begin{aligned}
O^{\mathrm{A}} & =\operatorname{Attn} \left(Q, A, \operatorname{Attn}^{\mathrm{S}}(A, K, V)\right) \\
& =\sigma\left(Q A^{T}\right) \sigma\left(A K^{T}\right) V
\end{aligned}
\end{eqnarray}
where $Q, K, V$ denote query, key and value matrices and $\sigma ( \cdot )$ represents Softmax function. Please refer to \cite{han2023agent} for more details. Although agent attention achieves a linear computation complexity of $O ( N n d )$, incorporating it in the optimization process leads to increased memory usage. Therefore, we will utilize it as an optional module when higher accuracy is required. We illustrate its effectiveness in ablation experiments.


\noindent{\bf T-GRU.} We build a T-GRU based on GRU blocks and our time embedding,it can be defined as follows:
\begin{eqnarray}
\begin{aligned}
z_{k} & =\sigma\left(\operatorname{Conv}\left(\left[h_{k-1}, x_{k}\right], W_{z}\right)+c_{k}+\mathbf{t}_{e}\right), \\
r_{k} & =\sigma\left(\operatorname{Conv}\left(\left[h_{k-1}, x_{k}\right], W_{r}\right)+c_{r}+\mathbf{t}_{e}\right), \\
\tilde{h}_{k} & =\tanh \left(\operatorname{Conv}\left(\left[r_{k} \odot h_{k-1}, x_{k}\right], W_{h}\right)+c_{h}+\mathbf{t}_{e}\right), \\
h_{k} & =\left(1-z_{k}\right) \odot h_{k-1}+z_{k} \odot \tilde{h}_{k},
\end{aligned}
\end{eqnarray}
where $h _ { k }$ is the hidden state; $c _ { k } , c _ { r } , c _ { h }$ are context features generated from the context network; $x_{k}$ is output features from BME or AA, and $W_{z} , W_{r} ,W_{h}$ are weight matrices that are learned. The number of channels in all features of ConvGRUs is 128.
 
\subsection{Loss Functions}
We calculate the Smooth L1 loss \cite{chang2018pyramid} on initial disparity $\mathbf{ d _ { 0 }}$ regressed from a group-wise correlation volume $\mathbf{C} _ { G } $,
\begin{eqnarray}
\mathcal{L}_ { i n i t } = S m o o t h _ { L _ { 1 } } (\mathbf{ d _ { 0 }} - \mathbf{d _ { g t }} )
\end{eqnarray}
where $\mathbf{d _ { g t }}$ represents the ground truth disparity. 

We calculate the L1 loss on all updated disparities $ \left\{ \mathbf{ d _ { i }} \right\} _ { i = 1 } ^ { N }$. The direct pixel-wise depth loss defined in Eq. 13.
\begin{eqnarray}
 \mathcal{L}_ {pixel}  =\sum_{i=1}^{N} \gamma^{N-i}\left\|\mathbf{d}_{i}  - \mathbf{d}_{gt}\right\|_{1}, \text { where } \gamma=0.9
\end{eqnarray}
where $N$ is the number of iterations, $\mathbf{ d _ {i}}$ is the disparity maps generated during inference process.

The diffusion loss $\mathcal{L} _ {diff}$ is defined as:
$$ \mathcal{L} _ {diff} =1- \frac { 1 } { M } \sum _ { j = 1 } ^ { M } S S I M (\mathbf{d}^{j}   , \mathbf{d}_{gt}^{j} )$$
where $SSIM$ is a quality assessment algorithm \cite{wang2004image}; $\mathbf{d}^{j}$ and $\mathbf{d}_{gt}^{j}$ are the image contents at the th local window; and $M$ is the number of local windows of the disp map $\mathbf{d}$. The total loss is defined as:
\begin{eqnarray}
\mathcal{L}=\lambda_{\mathrm{1}} \mathcal{L}_ { i n i t }+\mathcal{L}_ {pixel} +\lambda_{\text {2}} \mathcal{L} _ {diff}
\end{eqnarray}
where $\lambda_{\mathrm{1}}=1.0$ and $\lambda_{\mathrm{2}}=0.5$. DMIO is trained by combining losses through a weighted sum and minimizing the $\mathcal{L}$ defined.

\section{EXPERIMENTS}
\subsection{Datasets}
Scene Flow \cite{mayer2016large} is a synthetic dataset containing 35,454 training pairs and 4,370 testing pairs with dense disparity maps. We use the Finalpass of Scene Flow because it resembles real-world images more closely than the Cleanpass, which includes more motion blur and defocus. KITTI 2012 \cite{geiger2012we} and KITTI 2015 \cite{menze2015object} are datasets for real-world driving scenes. KITTI 2012 contains 194 training pairs and 195 testing pairs, while KITTI 2015 contains 200 training pairs and 200 testing pairs. Both datasets provide sparse ground-truth disparities obtained with LiDAR. Middlebury 2014 \cite{scharstein2014high} is an indoor dataset that includes 15 training pairs and 15 testing pairs. Some samples in the dataset are captured under inconsistent illumination or color conditions. All of the images are available in three different resolutions. ETH3D \cite{schops2017multi} is a grayscale dataset consisting of 27 training pairs and 20 testing pairs. We use the training pairs from Middlebury 2014 and ETH3D to evaluate cross-domain generalization performance.

\begin{figure}[!t]
	\centering
	\includegraphics[width=3.5in]{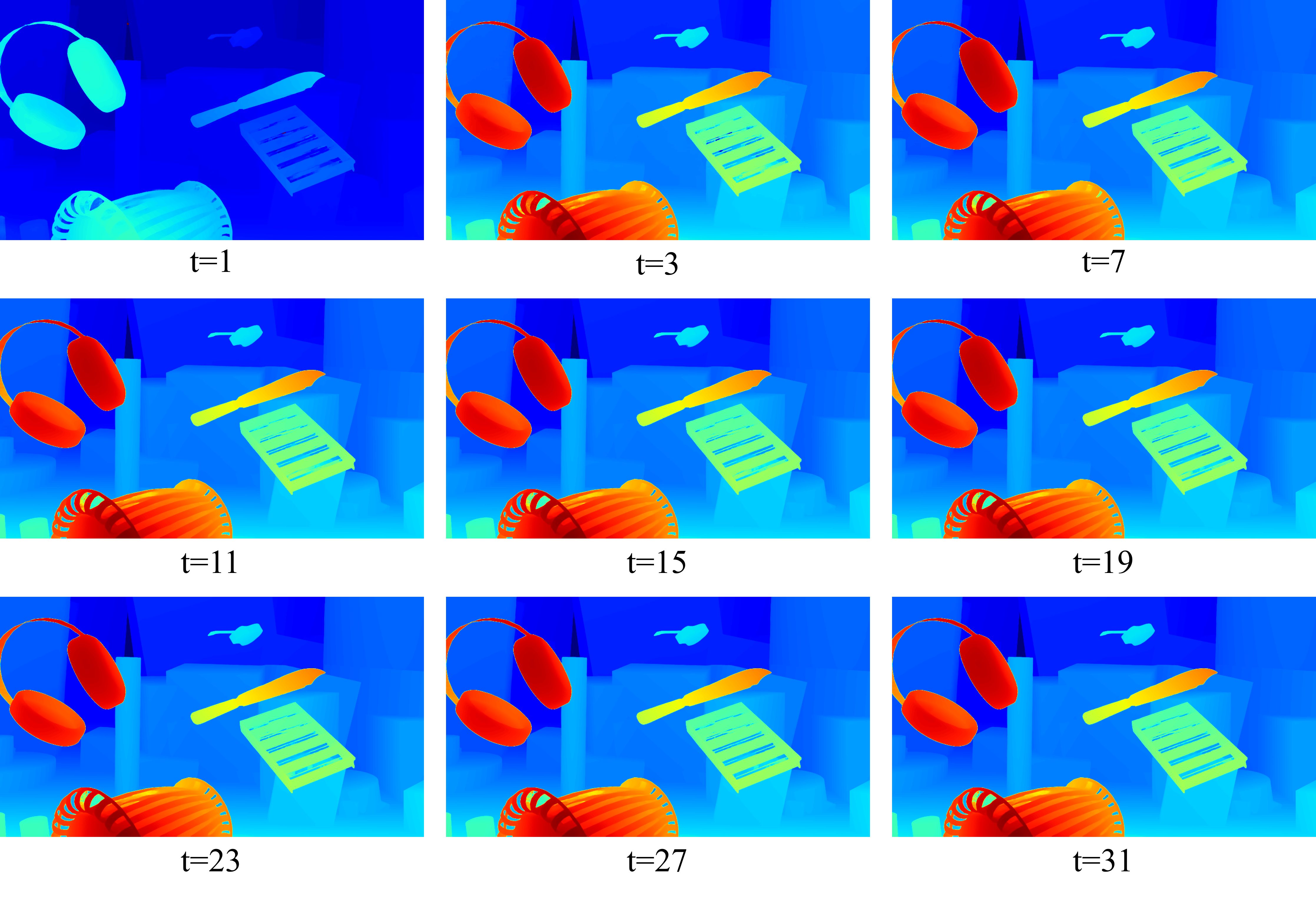}
	\caption{The visualization at different sampling steps on Scene Flow Dataset.t denotes the inference step.}
	\label{fig_1}
\end{figure}

\subsection{Implementation Details} We implemented our DMIO with PyTorch and conducted our experiments using NVIDIA V100 GPUs. For all training, we utilize the AdamW optimizer \cite{adamw} and clip gradients within the range of $\left [ -1, 1 \right ] $. On Scene Flow, we train DMIO for $400k$ steps with a batch size of $8$. On KITTI, we fine-tune the pre-trained Scene Flow model using a mixed dataset of KITTI 2012 and KITTI 2015 training image pairs for 50k steps. Ground truth annotations for KITTI are often sparse and noisy due to highly reflective surfaces, light-absorbing surfaces \cite{stommel2013inpainting}, dynamic objects \cite{menze2015joint}, and other factors. Therefore, as shown in Figure 5, we use nearest neighbor interpolation to fill in the missing ground truth values. We randomly crop images to $320 \times 736$ and apply the same data augmentation techniques as described in \cite{lipson2021raft} during training. The indexing radius is set to $4$. For all experiments, we utilize a one-cycle learning rate schedule with a learning rate of 0.0002, and we employ $22$ update iterations during training.

\subsection{Ablation Study}
In this section, we conducted an ablation study on the SceneFlow dataset, using the standard end-point error (EPE) and 3-pixel error (percentage of pixels with endpoint errors greater than 3px) of the overall regions as evaluation metrics. We use \cite{xu2023iterative} as the baseline and analyzed the effectiveness of various components in our approach.

\noindent{\bf Effectiveness of proposed modules.} We employ multiple strategies to enhance model performance. To verify the contribution of each strategy, we conducted experiments with various combinations and evaluated the performance of depth estimation. As shown in Table 1, each strategy enhances our network performance by several points. Without bells and whistles, our diffusion training method, which only involves TE, can improve accuracy. Our attempt to avoid iterating over the hidden state yielded poor results. As seen in Table 1, integrating a FFN into the contextual network does not result in a positive effect. Therefore, we only include the SMISH activation function after the CA layer. Due to the mathematically continuous nature of diffusion training, we need to incorporate additional training steps to enhance the results. The impact of algorithm AA is amplified by multiple iterations of optimization, leading to a 3.5\% improvement in performance compared to the method without AA. By combining all strategies, our method achieves highly promising performance in stereo matching tasks. The DMIO full module achieves a 7.52\% reduction in the EPE, which reduce the error of the base model from 0.47\% to 0.44\%.

\begin{table*}[htbp]
	\centering
	\scriptsize
	\caption{Ablation study of proposed networks on the Scene Flow test set. CA denotes Channel Self-Attention, FFN denotes Feed Forward Network, SMISH denotes $smish$ activation function, TE denotes Channel Self-Attention, AA denotes Agent Attention, Hide-free denotes Channel Self-Attention, and $400K$ denotes train steps. The baseline is our trained IGEV-Stereo.}
\begin{tabular}{cccccccc|cc}
	\toprule
	Baseline \cite{xu2023iterative} & CA & FFN  & SMISH & TE    & AA    & Hide-free & $400K$  & EPE (px) & $>$3px (\%)\\
	\midrule
	\midrule
	$\checkmark$     &       &       &       &       &       &       &       & 0.479 & 2.497 \\
	\midrule
	$\checkmark$     &       &       &       & $\checkmark$     &       & $\checkmark$     &       & 0.504 & 2.661 \\
	$\checkmark$     &       &       &       & $\checkmark$     &       &       &       & 0.470  & 2.452 \\
	$\checkmark$     & $\checkmark$     &       &       &       &       &       &       & 0.468 & 2.441 \\
	$\checkmark$     & $\checkmark$     & $\checkmark$     &       &       &       &       &       & 0.474 & 2.447 \\
	$\checkmark$     & $\checkmark$     &       & $\checkmark$     &       &       &       &       & 0.463 & 2.415 \\
	$\checkmark$     & $\checkmark$     &       & $\checkmark$     & $\checkmark$     &       &       & $\checkmark$     & 0.459 & 2.388 \\
	$\checkmark$     & $\checkmark$     &       & $\checkmark$     & $\checkmark$     & $\checkmark$     &       & $\checkmark$     & 0.443 & 2.293 \\
	\bottomrule
\end{tabular}%
	\label{tg}%
\end{table*}%

\begin{figure*}[t]
	\centering
	\includegraphics[width=16cm,height=4.91cm]{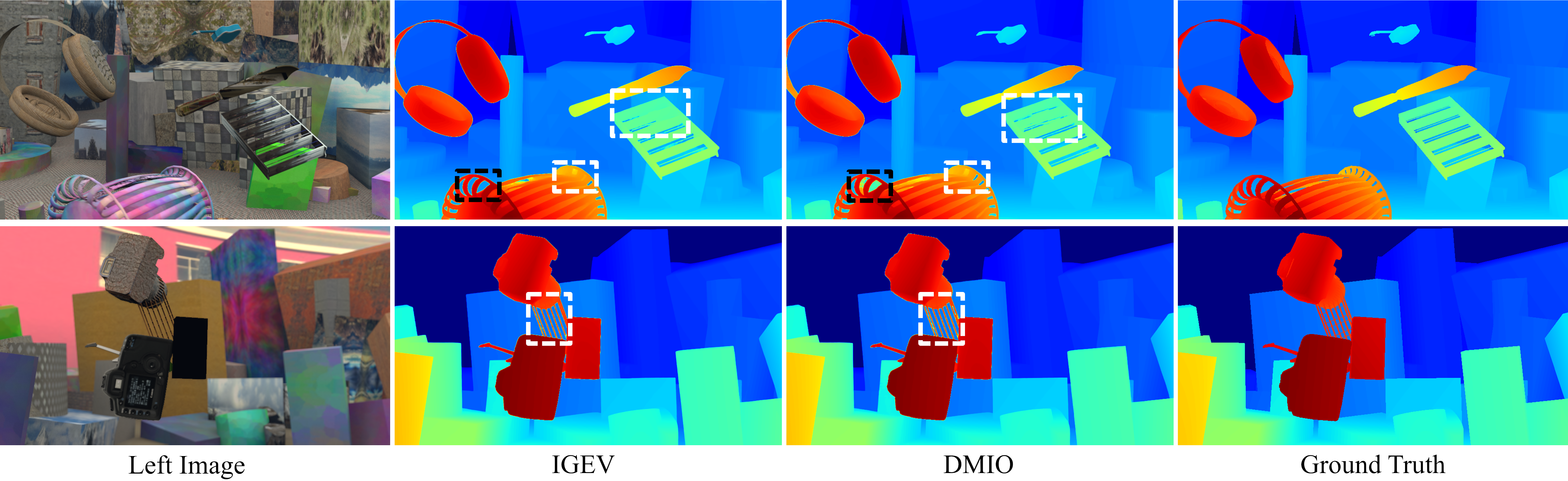}
	\caption{The visualization of DMIO compared with IGEV-Stereo on Scene Flow.}	
	\label{fig_6}
\end{figure*}

\noindent{\bf Number of Iterations.} To intuitively understand how the iterative optimization process refines the disparity step by step, we visualize the process as shown in Figure 6. We observe that the process initializes the disparity and achieves good results in only a few steps, then gradually improves the details, and does not over-refine the disparity map for large steps.

To further reveal the properties of using different inference steps, Table 2 presents the EPE of our DMIO under various settings. Compared to IGEV-Stereo, our method achieves better performance with the same number of iterations. Even when the number of iterations is reduced to 1, the full model only needs 8 iterations to achieve better performance. In practice, fewer inference steps are beneficial for reducing GPU memory consumption and speeding up inference.

\begin{table}[htbp]
	\centering
	\scriptsize
	\caption{ Ablation study for number of iterations.}
	\begin{tabular}{ccccccc}
		\toprule
		\multirow{2}[4]{*}{Model} & \multicolumn{6}{c}{Number of Iterations} \\
		\cmidrule{2-7}      & 1     & 2     & 3     & 4     & 8     & 32 \\
		\midrule
		Baseline \cite{xu2023iterative} & 0.66  & 0.62  & 0.58  & 0.55  & 0.50  & 0.47 \\
		Baseline+CA     & 0.66  & 0.61  & 0.57  & 0.64  & 0.49  & 0.46 \\
		Baseline+CA+TE & 0.65  & 0.60   & 0.56  & 0.53  & 0.48  & 0.45 \\
		Full model & 0.63  & 0.58  & 0.54  & 0.51  & 0.46  & 0.44 \\
		\bottomrule
	\end{tabular}%
	
	\label{tg}%
\end{table}%


\noindent{\bf Effectiveness of noises schedule.} This experiment shows the impact of sigmoid and linear noise schedules. As shown in Table 3, the sigmoid function requires adjustments to multiple hyperparameters to achieve optimal performance at 1/4 image resolution, and there is no specific rule to guide this process. To find the optimal schedule, we generally recommend using a simple linear schedule.

\begin{table}[htbp]
	\centering
	\scriptsize
	\caption{Ablation study for different noise schedule functions.}
\begin{tabular}{ccc}
	\toprule
	Noise Schedule Function & EPE (px) & $>$3px (\%) \\
	\midrule
	1 - t   & 0.478 & 2,533 \\
	\midrule
	sigmoid (s=0,e=3,T = 0.3) & 0.482 & 2.536 \\
	sigmoid (s=0,e=3,T = 0.7) & 0.479 & 2.513 \\
	sigmoid (s=-3,e=3,T = 1.0) & 0.480  & 2.517 \\
	sigmoid (s=-3,e=3,T = 1.1) & 0.470  & 2.452 \\
	\bottomrule
\end{tabular}%
	
	\label{tg}%
\end{table}%


\subsection{Performance Evaluation}In this subsection, we compare our method with other state-of-the-art methods using multiple datasets.

\begin{table*}[htbp]
	\centering
	\scriptsize
	\caption{Quantitative evaluation on Scene Flow test set. \textbf{Bold}: Best.}
	\begin{tabular}{ccccccccccc}
		\toprule
		Method & PSMNet & GANet & CSPN  & GwcNet & CStereo & CFNet & ACVNet & IGEV  & DiffuVolume & \textbf{Ours} \\
		EPE(px) & 1.09  & 0.84  & 0.78  & 0.76  & 0.72  & 0.70   & 0.48  & 0.47  & 0.46  & \textbf{0.44} \\
		\bottomrule
	\end{tabular}%
	\label{tg}%
\end{table*}%

\noindent{\bf Scene Flow.} The experiments in Table 4 show that we achieved state-of-the-art performance of 0.44 on the Scene Flow dataset. Compared to the classical PSMNet \cite{chang2018pyramid}, our DMIO achieves close to 2.5 times better accuracy and also outperforms the results of the DiffuVolume \cite{zheng2023diffuvolume} work from the same period. In Figure 7, we further compare our DMIO results with the excellent IGEV in randomly select images. The prediction quality of DMIO in the edge region outperforms that of IGEV. It is worth mentioning that in specific edge regions, DMIO closely approximates the accuracy of the ground truth. The black box in Figure 7 highlights the failure of DMIO. While DMIO may yield better numerical values in EPE, not all pixels exhibit improved results, indicating shortcomings in capturing details at close range.

\begin{figure*}[t]
	\centering
	\includegraphics[width=16cm,height=5.837cm]{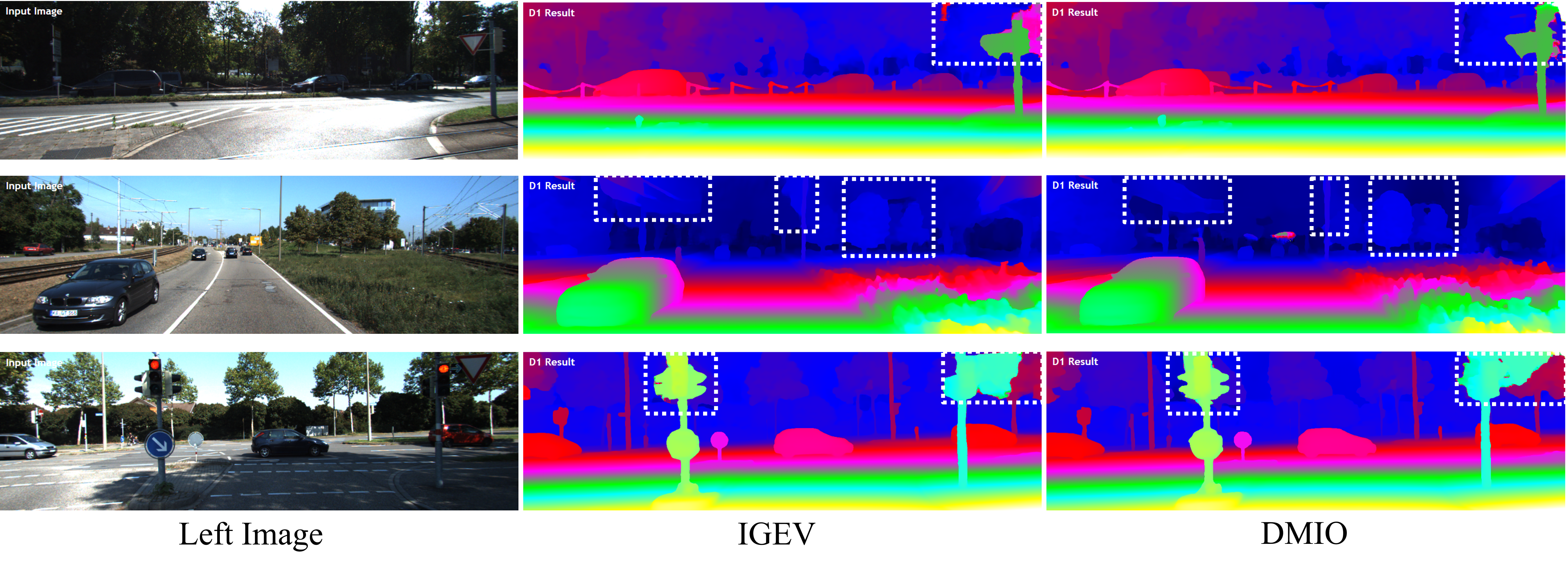}
	\caption{Qualitative results on the test set of KITTI. Our DMIO outperforms IGEV in detailed and weak texture regions.}	
	\label{fig_6}
\end{figure*}

\begin{table*}[htbp]
	\centering
	\scriptsize
	\caption{Synthetic to real generalization experiments. All models are trained on Scene Flow. The 2-pixel error rate is used for Middlebury 2014, and 1-pixel error rate for ETH3D.
	} 
	\begin{tabular}{c|cccccc|ccc}
		\toprule     
		\multirow{2}[2]{*}{Method} & \multicolumn{6}{c|}{KITTI 2012 }          & \multicolumn{3}{c}{KITTI 2015 } \\  
		& 2-noc & 2-all & 3-noc & 3-all & EPE noc & EPE  all & D1-bg & D1-fg & D1-all \\
		\midrule 
		PSMNet \cite{chang2018pyramid} & 2.44  & 3.01  & 1.49  & 1.89  & 0.5   & 0.6   & 1.86  & 4.62  & 2.32 \\
		GwcNet \cite{guo2019group} & 2.16  & 2.71  & 1.32  & 1.70  & 0.5   & 0.5   & 1.74  & 3.93  & 2.11 \\
		GANet-deep \cite{zhang2019ga} & 1.89  & 2.50  & 1.19  & 1.60  & 0.4   & 0.5   & 1.48  & 3.46  & 1.81 \\
		AcfNet \cite{zhang2020adaptive} & 1.83  & 2.35  & 1.17  & 1.54  & 0.5   & 0.5   & 1.51  & 3.80  & 1.89 \\
		HITNet \cite{tankovich2021hitnet} & 2.00  & 2.65  & 1.41  & 1.89  & 0.4   & 0.5   & 1.74  & 3.20  & 1.98 \\
		EdgeStereo-V2 \cite{song2020edgestereo} & 2.32  & 2.88  & 1.46  & 1.83  & 0.4   & 0.5   & 1.84  & 3.30  & 2.08 \\
		CSPN \cite{shen2021cfnet} & 1.79  & 2.27  & 1.19  & 1.53  &   -   &   -  & 1.51  & 2.88  & 1.74 \\
		LEAStereo \cite{song2020edgestereo} & 1.90  & 2.39  & 1.13  & 1.45  & 0.5   & 0.5   & 1.40  & 2.91  & 1.65 \\  
		ACVNet \cite{xu2022attention} & 1.83  & 2.35  & 1.13  & 1.47  & 0.4   & 0.5   & \textbf{1.37}  & 3.07  & 1.65 \\ 
		CREStereo \cite{li2022practical} & 1.72  & 2.18  & 1.14  & 1.46  & 0.4   & 0.5   & 1.45  & 2.86  & 1.69 \\
		RAFT-Stereo \cite{lipson2021raft} & 1.92  & 2.42  & 1.30  & 1.66  & 0.4   & 0.5   & 1.58  & 3.05  & 1.82 \\
		IGEV-Stereo \cite{xu2023iterative}   & \textbf{1.71}  & \textbf{2.17}  & \textbf{1.12}  & \textbf{1.44}  & 0.4   & 0.4   & 1.38  &  2.67  & \textbf{1.59} \\
		DMIO (Ours) & 1.73  & 2.20  & 1.14  & 1.48  & 0.4   & 0.4   & 1.45  & \textbf{2.61}  & 1.64 \\
		\bottomrule
	\end{tabular}%
	\label{tg}%
\end{table*}%

\noindent{\bf KITTI.} For the evaluation of the KITTI dataset, we adhere to the standard protocol by submitting our fine-tuned results to the KITTI leaderboard [16]. Table 5 demonstrates our method compared to previous state-of-the-art methods. We evaluated the performance on the KITTI 2012 and 2015 test sets. In our assessment of various regions, our method excels notably in its performance on foreground objects, such as cars and pedestrians, achieving a D1-fg score of 2.61\%. The results surpass very recent methods, including RAFT-Stereo and IGEV-Stereo. Unfortunately, we only achieve the best results in D1-fg, but there is only a very small difference with IGEV in other metrics. We rank second in the most important D1-all metric.

As shown in Figure 7, for out-of-field regions marked by white boxes that lack corresponding pixels, our proposed DMIO can still successfully estimate disparity, demonstrating better depth consistency and clearer structure. Notably, the KITTI dataset lacks LiDAR ground truth and dense annotations on the upper part of the image, as shown in Figure 5. Additionally, portions of these results are not evaluated in the D1-all error. This lack of annotation may introduce bias in the final D1-all error, preventing the full revelation of network effectiveness. Figure 7 illustrates this challenge by comparing DMIO with the state-of-the-art IGEVStereo. Even though DMIO produces larger D1-all errors, the visualization results show significantly better structure and fewer artifacts in regions lacking true disparity.

As shown in Table 6, we assess the performance of IGEV-Stereo and DMIO in reflective regions Although DMIO lags behind IGEV in the Test Set Average index of KITTI 2012, it achieves better results in 2-pixels and 5-pixels in 32 iterations. In general, DMIO and IGEV achieve similar results in the KITTI benchmark.

\subsection{Zero-Shot Generalization} Given the challenge of acquiring large real-world datasets for training, the ability of stereo models to generalize is crucial. We evaluate the generalization performance of DMIO from synthetic datasets to unseen real-world scenes. In this evaluation, we trained our DMIO on Scene Flow using data augmentation and directly tested it on the Middlebury 2014 and ETH3D training sets. As shown in the table. In our study, our DMIO achieves state-of-the-art performance in the same zero-shot setting. Fig. 9 shows a visual comparison with IGEV-Stereo; our method is more robust for textureless and detailed regions.

\begin{figure*}[t]
	\centering
	\includegraphics[width=16cm,height=11.79cm]{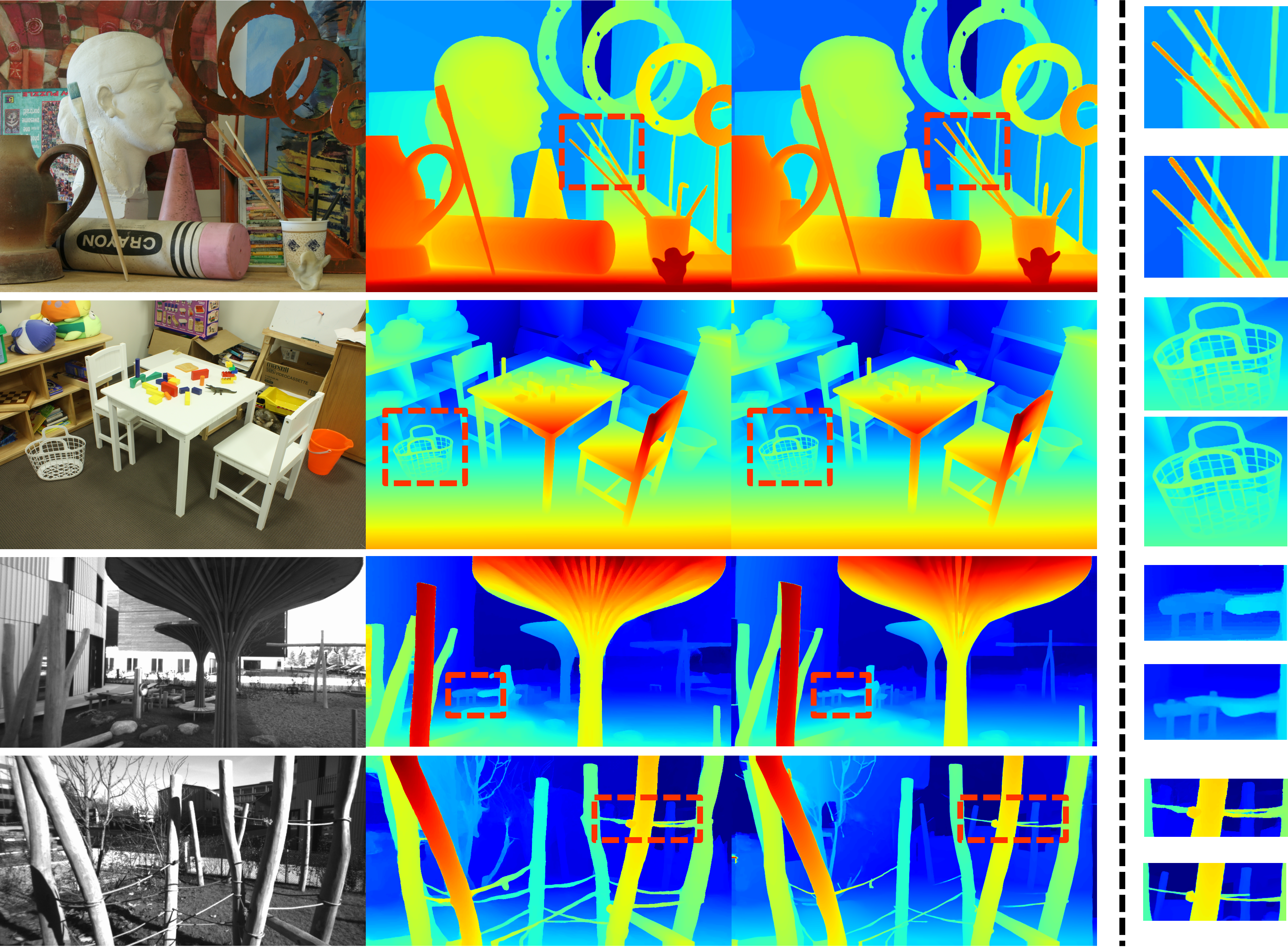}
	\caption{Generalization results on Middlebury 2014 and ETH3D. The second and third columns display the results of IGEV-Stereo and our method, respectively.}	
	\label{fig_6}
\end{figure*}

\begin{table}[htbp]
	\centering
	\scriptsize
	\caption{Evaluation in the reflective regions (ill-posed regions) of KITTI 2012 benchmark. Iter. denotes iteration number.
	}
	\begin{tabular}{c|c|c|cc}
		\toprule
		Method & Iter. & pixels & noc   & all \\
		\midrule
		\multirow{4}[2]{*}{IGEV-Stereo} & \multirow{4}[2]{*}{32} & 2     & 7.29  & 8.48 \\
		&       & 3     & \textbf{4.11}  & \textbf{4.76} \\
		&       & 4     & \textbf{2.92}  & \textbf{3.35} \\
		&       & 5     & 2.34  & 2.68 \\
		\midrule
		\multirow{4}[2]{*}{DMIO} & \multirow{4}[2]{*}{32} & 2     & \textbf{7.23}  & \textbf{8.43} \\
		&       & 3     & 4.21  & 4.89 \\
		&       & 4     & 2.97  & 3.43 \\
		&       & 5     & \textbf{2.29}  & \textbf{2.63} \\
		\bottomrule
	\end{tabular}%
	\label{tg}%
\end{table}%

\begin{table}[htbp]
	\centering
	\scriptsize
	\caption{Synthetic to real generalization experiments. All models are trained on Scene Flow. The 2-pixel error rate is used for Middlebury 2014, and 1-pixel error rate for ETH3D.
}
\begin{tabular}{cccc}
	\toprule
	\multirow{2}[2]{*}{Model} & \multicolumn{2}{c}{Middlebury} & \multirow{2}[2]{*}{ETH3D} \\
	& half  & quarter &  \\
	\midrule
	CostFilter [19] & 40.5  & 17.6  & 31.1 \\
	PatchMatch [3] & 38.6  & 16.1  & 24.1 \\
	SGM [18] & 25.2  & 10.7  & 12.9 \\
	\midrule
	PSMNet [5] & 15.8  & 9.8   & 10.2 \\
	GANet [56] & 13.5  & 8.5   & 6.5 \\
	DSMNet [57] & 13.8  & 8.1   & 6.2 \\
	STTR [22] & 15.5  & 9.7   & 17.2 \\
	CFNet [36] & 15.3  & 9.8   & 5.8 \\
	FC-GANet [58] & 10.2  & 7.8   & 5.8 \\
	Graft-GANet [26] & 9.8   &       & 6.2 \\
	RAFT-Stereo [24] & 8.7   & 7.3   & \textbf{3.2} \\
	IGEV-Stereo & 7.1   & 6.2   & 3.6 \\
	DMIO (Ours) & \textbf{6.1} & \textbf{5.8} & 3.8 \\
	\bottomrule
\end{tabular}%
\label{tg}%
\end{table}%


\section{Conclusion and Future Work} In this paper, we propose a novel diffusion model-based stereo matching network, called DMIO, which utilizes the attention mechanism and the diffusion model to iteratively optimize the coarse disparity. An attention-based context network is proposed, utilizing a linear channel attention mechanism to extract more refined context features. This approach enhances the performance of dispariy estimation and offers a solid initial value for subsequent dispariy optimization. We propose a Time-based Gated Recurrent Unit to help the network capture detailed information about edges and smooth regions. Additionally, we embed the diffusion model into the optimization module by utilizing Time encoding and Agent Attention. As we conclude this paper, our method outperforms recent state-of-the-art methods on the SceneFlow dataset and achieves competitive results on the KITTI image leaderboard.

However, our approach still faces some challenges. Firstly, limited by the diffusion model, our method requires dense ground truth. Therefore, we will investigate combining more advanced completion methods with our approach to enhance accuracy. Secondly, the utilization of 3D CNN and attention mechanism in this paper results in high computational and memory costs. Future research will focus on designing a more lightweight network. Finally, with the rapid development of diffusion models, we will explore the application of iterative optimization modules for the latest diffusion methods, aiming to reduce the number of iterations.



\section*{Acknowledgments}
This work was supported by the National Natural Science Foundation of China (No. 62271143), and the Big Data Computing Center of Southeast University. 

\bibliography{tip}{}

\begin{thebibliography}{10}
\providecommand{\url}[1]{#1}
\csname url@samestyle\endcsname
\providecommand{\newblock}{\relax}
\providecommand{\bibinfo}[2]{#2}
\providecommand{\BIBentrySTDinterwordspacing}{\spaceskip=0pt\relax}
\providecommand{\BIBentryALTinterwordstretchfactor}{4}
\providecommand{\BIBentryALTinterwordspacing}{\spaceskip=\fontdimen2\font plus
\BIBentryALTinterwordstretchfactor\fontdimen3\font minus
  \fontdimen4\font\relax}
\providecommand{\BIBforeignlanguage}[2]{{%
\expandafter\ifx\csname l@#1\endcsname\relax
\typeout{** WARNING: IEEEtran.bst: No hyphenation pattern has been}%
\typeout{** loaded for the language `#1'. Using the pattern for}%
\typeout{** the default language instead.}%
\else
\language=\csname l@#1\endcsname
\fi
#2}}
\providecommand{\BIBdecl}{\relax}
\BIBdecl

\bibitem{zhao2023high}
H.~Zhao, H.~Zhou, Y.~Zhang, J.~Chen, Y.~Yang, and Y.~Zhao, ``High-frequency
  stereo matching network,'' in \emph{Proceedings of the IEEE/CVF Conference on
  Computer Vision and Pattern Recognition}, 2023, pp. 1327--1336.

\bibitem{li2023bbdm}
B.~Li, K.~Xue, B.~Liu, and Y.-K. Lai, ``Bbdm: Image-to-image translation with
  brownian bridge diffusion models,'' in \emph{Proceedings of the IEEE/CVF
  Conference on Computer Vision and Pattern Recognition}, 2023, pp. 1952--1961.

\bibitem{liu2022flow}
X.~Liu, C.~Gong, and Q.~Liu, ``Flow straight and fast: Learning to generate and
  transfer data with rectified flow,'' \emph{arXiv preprint arXiv:2209.03003},
  2022.

\bibitem{lipson2021raft}
L.~Lipson, Z.~Teed, and J.~Deng, ``Raft-stereo: Multilevel recurrent field
  transforms for stereo matching,'' in \emph{2021 International Conference on
  3D Vision (3DV)}.\hskip 1em plus 0.5em minus 0.4em\relax IEEE, 2021, pp.
  218--227.

\bibitem{zhang2020adaptive}
Y.~Zhang, Y.~Chen, X.~Bai, S.~Yu, K.~Yu, Z.~Li, and K.~Yang, ``Adaptive
  unimodal cost volume filtering for deep stereo matching,'' in
  \emph{Proceedings of the AAAI Conference on Artificial Intelligence},
  vol.~34, no.~07, 2020, pp. 12\,926--12\,934.

\bibitem{kendall2017end}
A.~Kendall, H.~Martirosyan, S.~Dasgupta, P.~Henry, R.~Kennedy, A.~Bachrach, and
  A.~Bry, ``End-to-end learning of geometry and context for deep stereo
  regression,'' in \emph{Proceedings of the IEEE international conference on
  computer vision}, 2017, pp. 66--75.

\bibitem{chang2018pyramid}
J.-R. Chang and Y.-S. Chen, ``Pyramid stereo matching network,'' in
  \emph{Proceedings of the IEEE conference on computer vision and pattern
  recognition}, 2018, pp. 5410--5418.

\bibitem{duggal2019deeppruner}
S.~Duggal, S.~Wang, W.-C. Ma, R.~Hu, and R.~Urtasun, ``Deeppruner: Learning
  efficient stereo matching via differentiable patchmatch,'' in
  \emph{Proceedings of the IEEE/CVF international conference on computer
  vision}, 2019, pp. 4384--4393.

\bibitem{cheng2020hierarchical}
X.~Cheng, Y.~Zhong, M.~Harandi, Y.~Dai, X.~Chang, H.~Li, T.~Drummond, and
  Z.~Ge, ``Hierarchical neural architecture search for deep stereo matching,''
  \emph{Advances in Neural Information Processing Systems}, vol.~33, pp.
  22\,158--22\,169, 2020.

\bibitem{wang2021pvstereo}
H.~Wang, R.~Fan, P.~Cai, and M.~Liu, ``Pvstereo: Pyramid voting module for
  end-to-end self-supervised stereo matching,'' \emph{IEEE Robotics and
  Automation Letters}, vol.~6, no.~3, pp. 4353--4360, 2021.

\bibitem{shen2022pcw}
Z.~Shen, Y.~Dai, X.~Song, Z.~Rao, D.~Zhou, and L.~Zhang, ``Pcw-net: Pyramid
  combination and warping cost volume for stereo matching,'' in \emph{European
  Conference on Computer Vision}.\hskip 1em plus 0.5em minus 0.4em\relax
  Springer, 2022, pp. 280--297.

\bibitem{guo2019group}
X.~Guo, K.~Yang, W.~Yang, X.~Wang, and H.~Li, ``Group-wise correlation stereo
  network,'' in \emph{Proceedings of the IEEE/CVF conference on computer vision
  and pattern recognition}, 2019, pp. 3273--3282.

\bibitem{zhang2019ga}
F.~Zhang, V.~Prisacariu, R.~Yang, and P.~H. Torr, ``Ga-net: Guided aggregation
  net for end-to-end stereo matching,'' in \emph{Proceedings of the IEEE/CVF
  Conference on Computer Vision and Pattern Recognition}, 2019, pp. 185--194.

\bibitem{xu2022attention}
G.~Xu, J.~Cheng, P.~Guo, and X.~Yang, ``Attention concatenation volume for
  accurate and efficient stereo matching,'' in \emph{Proceedings of the
  IEEE/CVF Conference on Computer Vision and Pattern Recognition}, 2022, pp.
  12\,981--12\,990.

\bibitem{cheng2020deep}
S.~Cheng, Z.~Xu, S.~Zhu, Z.~Li, L.~E. Li, R.~Ramamoorthi, and H.~Su, ``Deep
  stereo using adaptive thin volume representation with uncertainty
  awareness,'' in \emph{Proceedings of the IEEE/CVF Conference on Computer
  Vision and Pattern Recognition}, 2020, pp. 2524--2534.

\bibitem{yang2020cost}
J.~Yang, W.~Mao, J.~M. Alvarez, and M.~Liu, ``Cost volume pyramid based depth
  inference for multi-view stereo,'' in \emph{Proceedings of the IEEE/CVF
  Conference on Computer Vision and Pattern Recognition}, 2020, pp. 4877--4886.

\bibitem{gu2020cascade}
X.~Gu, Z.~Fan, S.~Zhu, Z.~Dai, F.~Tan, and P.~Tan, ``Cascade cost volume for
  high-resolution multi-view stereo and stereo matching,'' in \emph{Proceedings
  of the IEEE/CVF conference on computer vision and pattern recognition}, 2020,
  pp. 2495--2504.

\bibitem{mao2021uasnet}
Y.~Mao, Z.~Liu, W.~Li, Y.~Dai, Q.~Wang, Y.-T. Kim, and H.-S. Lee, ``Uasnet:
  Uncertainty adaptive sampling network for deep stereo matching,'' in
  \emph{Proceedings of the IEEE/CVF International Conference on Computer
  Vision}, 2021, pp. 6311--6319.

\bibitem{khamis2018stereonet}
S.~Khamis, S.~Fanello, C.~Rhemann, A.~Kowdle, J.~Valentin, and S.~Izadi,
  ``Stereonet: Guided hierarchical refinement for edge-aware depth
  prediction,'' 2018.

\bibitem{zhang2018activestereonet}
Y.~Zhang, S.~Khamis, C.~Rhemann, J.~Valentin, A.~Kowdle, V.~Tankovich,
  M.~Schoenberg, S.~Izadi, T.~Funkhouser, and S.~Fanello, ``Activestereonet:
  End-to-end self-supervised learning for active stereo systems,'' in
  \emph{Proceedings of the european conference on computer vision (ECCV)},
  2018, pp. 784--801.

\bibitem{xu2021bilateral}
B.~Xu, Y.~Xu, X.~Yang, W.~Jia, and Y.~Guo, ``Bilateral grid learning for stereo
  matching networks,'' in \emph{Proceedings of the IEEE/CVF Conference on
  Computer Vision and Pattern Recognition}, 2021, pp. 12\,497--12\,506.

\bibitem{tankovich2021hitnet}
V.~Tankovich, C.~Hane, Y.~Zhang, A.~Kowdle, S.~Fanello, and S.~Bouaziz,
  ``Hitnet: Hierarchical iterative tile refinement network for real-time stereo
  matching,'' in \emph{Proceedings of the IEEE/CVF Conference on Computer
  Vision and Pattern Recognition}, 2021, pp. 14\,362--14\,372.

\bibitem{xu2023accurate}
G.~Xu, Y.~Wang, J.~Cheng, J.~Tang, and X.~Yang, ``Accurate and efficient stereo
  matching via attention concatenation volume,'' \emph{IEEE Transactions on
  Pattern Analysis and Machine Intelligence}, 2023.

\bibitem{li2021revisiting}
Z.~Li, X.~Liu, N.~Drenkow, A.~Ding, F.~X. Creighton, R.~H. Taylor, and
  M.~Unberath, ``Revisiting stereo depth estimation from a sequence-to-sequence
  perspective with transformers,'' in \emph{Proceedings of the IEEE/CVF
  international conference on computer vision}, 2021, pp. 6197--6206.

\bibitem{guo2022context}
W.~Guo, Z.~Li, Y.~Yang, Z.~Wang, R.~H. Taylor, M.~Unberath, A.~Yuille, and
  Y.~Li, ``Context-enhanced stereo transformer,'' in \emph{European Conference
  on Computer Vision}.\hskip 1em plus 0.5em minus 0.4em\relax Springer, 2022,
  pp. 263--279.

\bibitem{lou2023elfnet}
J.~Lou, W.~Liu, Z.~Chen, F.~Liu, and J.~Cheng, ``Elfnet: Evidential
  local-global fusion for stereo matching,'' in \emph{Proceedings of the
  IEEE/CVF International Conference on Computer Vision}, 2023, pp.
  17\,784--17\,793.

\bibitem{teed2020raft}
Z.~Teed and J.~Deng, ``Raft: Recurrent all-pairs field transforms for optical
  flow,'' in \emph{Computer Vision--ECCV 2020: 16th European Conference,
  Glasgow, UK, August 23--28, 2020, Proceedings, Part II 16}.\hskip 1em plus
  0.5em minus 0.4em\relax Springer, 2020, pp. 402--419.

\bibitem{xu2023iterative}
G.~Xu, X.~Wang, X.~Ding, and X.~Yang, ``Iterative geometry encoding volume for
  stereo matching,'' in \emph{Proceedings of the IEEE/CVF Conference on
  Computer Vision and Pattern Recognition}, 2023, pp. 21\,919--21\,928.

\bibitem{li2022practical}
J.~Li, P.~Wang, P.~Xiong, T.~Cai, Z.~Yan, L.~Yang, J.~Liu, H.~Fan, and S.~Liu,
  ``Practical stereo matching via cascaded recurrent network with adaptive
  correlation,'' in \emph{Proceedings of the IEEE/CVF conference on computer
  vision and pattern recognition}, 2022, pp. 16\,263--16\,272.

\bibitem{liu2024global}
Z.~Liu, Y.~Li, and M.~Okutomi, ``Global occlusion-aware transformer for robust
  stereo matching,'' in \emph{Proceedings of the IEEE/CVF Winter Conference on
  Applications of Computer Vision}, 2024, pp. 3535--3544.

\bibitem{jing2023uncertainty}
J.~Jing, J.~Li, P.~Xiong, J.~Liu, S.~Liu, Y.~Guo, X.~Deng, M.~Xu, L.~Jiang, and
  L.~Sigal, ``Uncertainty guided adaptive warping for robust and efficient
  stereo matching,'' in \emph{Proceedings of the IEEE/CVF International
  Conference on Computer Vision}, 2023, pp. 3318--3327.

\bibitem{karaev2023dynamicstereo}
N.~Karaev, I.~Rocco, B.~Graham, N.~Neverova, A.~Vedaldi, and C.~Rupprecht,
  ``Dynamicstereo: Consistent dynamic depth from stereo videos,'' in
  \emph{Proceedings of the IEEE/CVF Conference on Computer Vision and Pattern
  Recognition}, 2023, pp. 13\,229--13\,239.

\bibitem{chen2023diffusiondet}
S.~Chen, P.~Sun, Y.~Song, and P.~Luo, ``Diffusiondet: Diffusion model for
  object detection,'' in \emph{Proceedings of the IEEE/CVF International
  Conference on Computer Vision}, 2023, pp. 19\,830--19\,843.

\bibitem{amit2021segdiff}
T.~Amit, E.~Nachmani, T.~Shaharbany, and L.~Wolf, ``Segdiff: Image segmentation
  with diffusion probabilistic models,'' \emph{arXiv preprint
  arXiv:2112.00390}, 2021.

\bibitem{baranchuk2021label}
D.~Baranchuk, I.~Rubachev, A.~Voynov, V.~Khrulkov, and A.~Babenko,
  ``Label-efficient semantic segmentation with diffusion models,'' \emph{arXiv
  preprint arXiv:2112.03126}, 2021.

\bibitem{choi2021ilvr}
J.~Choi, S.~Kim, Y.~Jeong, Y.~Gwon, and S.~Yoon, ``Ilvr: Conditioning method
  for denoising diffusion probabilistic models,'' \emph{arXiv preprint
  arXiv:2108.02938}, 2021.

\bibitem{kawar2022denoising}
B.~Kawar, M.~Elad, S.~Ermon, and J.~Song, ``Denoising diffusion restoration
  models,'' \emph{Advances in Neural Information Processing Systems}, vol.~35,
  pp. 23\,593--23\,606, 2022.

\bibitem{saharia2022image}
C.~Saharia, J.~Ho, W.~Chan, T.~Salimans, D.~J. Fleet, and M.~Norouzi, ``Image
  super-resolution via iterative refinement,'' \emph{IEEE Transactions on
  Pattern Analysis and Machine Intelligence}, 2022.

\bibitem{lin2023diffbir}
X.~Lin, J.~He, Z.~Chen, Z.~Lyu, B.~Fei, B.~Dai, W.~Ouyang, Y.~Qiao, and
  C.~Dong, ``Diffbir: Towards blind image restoration with generative diffusion
  prior,'' \emph{arXiv preprint arXiv:2308.15070}, 2023.

\bibitem{meng2021sdedit}
C.~Meng, Y.~Song, J.~Song, J.~Wu, J.-Y. Zhu, and S.~Ermon, ``Sdedit: Image
  synthesis and editing with stochastic differential equations,'' \emph{arXiv
  preprint arXiv:2108.01073}, 2021.

\bibitem{ho2022video}
J.~Ho, T.~Salimans, A.~Gritsenko, W.~Chan, M.~Norouzi, and D.~J. Fleet, ``Video
  diffusion models,'' \emph{Advances in Neural Information Processing Systems},
  vol.~35, pp. 8633--8646, 2022.

\bibitem{duan2023diffusiondepth}
Y.~Duan, X.~Guo, and Z.~Zhu, ``Diffusiondepth: Diffusion denoising approach for
  monocular depth estimation,'' \emph{arXiv preprint arXiv:2303.05021}, 2023.

\bibitem{saxena2023monocular}
S.~Saxena, A.~Kar, M.~Norouzi, and D.~J. Fleet, ``Monocular depth estimation
  using diffusion models,'' \emph{arXiv preprint arXiv:2302.14816}, 2023.

\bibitem{ke2023repurposing}
B.~Ke, A.~Obukhov, S.~Huang, N.~Metzger, R.~C. Daudt, and K.~Schindler,
  ``Repurposing diffusion-based image generators for monocular depth
  estimation,'' \emph{arXiv preprint arXiv:2312.02145}, 2023.

\bibitem{saxena2023zero}
S.~Saxena, J.~Hur, C.~Herrmann, D.~Sun, and D.~J. Fleet, ``Zero-shot metric
  depth with a field-of-view conditioned diffusion model,'' \emph{arXiv
  preprint arXiv:2312.13252}, 2023.

\bibitem{shao2023monodiffusion}
S.~Shao, Z.~Pei, W.~Chen, D.~Sun, P.~C. Chen, and Z.~Li, ``Monodiffusion:
  Self-supervised monocular depth estimation using diffusion model,''
  \emph{arXiv preprint arXiv:2311.07198}, 2023.

\bibitem{saxena2024surprising}
S.~Saxena, C.~Herrmann, J.~Hur, A.~Kar, M.~Norouzi, D.~Sun, and D.~J. Fleet,
  ``The surprising effectiveness of diffusion models for optical flow and
  monocular depth estimation,'' \emph{Advances in Neural Information Processing
  Systems}, vol.~36, 2024.

\bibitem{resnet}
K.~He, X.~Zhang, S.~Ren, and J.~Sun, ``Deep residual learning for image
  recognition,'' in \emph{Proceedings of the IEEE conference on computer vision
  and pattern recognition}, 2016, pp. 770--778.

\bibitem{jiang2021learning}
S.~Jiang, D.~Campbell, Y.~Lu, H.~Li, and R.~Hartley, ``Learning to estimate
  hidden motions with global motion aggregation,'' in \emph{Proceedings of the
  IEEE/CVF international conference on computer vision}, 2021, pp. 9772--9781.

\bibitem{wang2022smish}
X.~Wang, H.~Ren, and A.~Wang, ``Smish: A novel activation function for deep
  learning methods,'' \emph{Electronics}, vol.~11, no.~4, p. 540, 2022.

\bibitem{zamir2022restormer}
S.~W. Zamir, A.~Arora, S.~Khan, M.~Hayat, F.~S. Khan, and M.-H. Yang,
  ``Restormer: Efficient transformer for high-resolution image restoration,''
  in \emph{Proceedings of the IEEE/CVF conference on computer vision and
  pattern recognition}, 2022, pp. 5728--5739.

\bibitem{ho2020denoising}
J.~Ho, A.~Jain, and P.~Abbeel, ``Denoising diffusion probabilistic models,''
  \emph{Advances in Neural Information Processing Systems}, vol.~33, pp.
  6840--6851, 2020.

\bibitem{zhou2023denoising}
L.~Zhou, A.~Lou, S.~Khanna, and S.~Ermon, ``Denoising diffusion bridge
  models,'' \emph{arXiv preprint arXiv:2309.16948}, 2023.

\bibitem{liu2023instaflow}
X.~Liu, X.~Zhang, J.~Ma, J.~Peng, and Q.~Liu, ``Instaflow: One step is enough
  for high-quality diffusion-based text-to-image generation,'' \emph{arXiv
  preprint arXiv:2309.06380}, 2023.

\bibitem{nichol2021improved}
A.~Q. Nichol and P.~Dhariwal, ``Improved denoising diffusion probabilistic
  models,'' in \emph{International Conference on Machine Learning}.\hskip 1em
  plus 0.5em minus 0.4em\relax PMLR, 2021, pp. 8162--8171.

\bibitem{chen2023importance}
T.~Chen, ``On the importance of noise scheduling for diffusion models,''
  \emph{arXiv preprint arXiv:2301.10972}, 2023.

\bibitem{j2022scalable}
A.~Jabri, D.~Fleet, and T.~Chen, ``Scalable adaptive computation for iterative
  generation,'' \emph{arXiv preprint arXiv:2212.11972}, 2022.

\bibitem{ronneberger2015u}
O.~Ronneberger, P.~Fischer, and T.~Brox, ``U-net: Convolutional networks for
  biomedical image segmentation,'' in \emph{Medical image computing and
  computer-assisted intervention--MICCAI 2015: 18th international conference,
  Munich, Germany, October 5-9, 2015, proceedings, part III 18}.\hskip 1em plus
  0.5em minus 0.4em\relax Springer, 2015, pp. 234--241.

\bibitem{si2023freeu}
C.~Si, Z.~Huang, Y.~Jiang, and Z.~Liu, ``Freeu: Free lunch in diffusion
  u-net,'' \emph{arXiv preprint arXiv:2309.11497}, 2023.

\bibitem{williams2024unified}
C.~Williams, F.~Falck, G.~Deligiannidis, C.~C. Holmes, A.~Doucet, and S.~Syed,
  ``A unified framework for u-net design and analysis,'' \emph{Advances in
  Neural Information Processing Systems}, vol.~36, 2024.

\bibitem{bao2023all}
F.~Bao, S.~Nie, K.~Xue, Y.~Cao, C.~Li, H.~Su, and J.~Zhu, ``All are worth
  words: A vit backbone for diffusion models,'' in \emph{Proceedings of the
  IEEE/CVF Conference on Computer Vision and Pattern Recognition}, 2023, pp.
  22\,669--22\,679.

\bibitem{peebles2023scalable}
W.~Peebles and S.~Xie, ``Scalable diffusion models with transformers,'' in
  \emph{Proceedings of the IEEE/CVF International Conference on Computer
  Vision}, 2023, pp. 4195--4205.

\bibitem{vaswani2017attention}
A.~Vaswani, N.~Shazeer, N.~Parmar, J.~Uszkoreit, L.~Jones, A.~N. Gomez,
  {\L}.~Kaiser, and I.~Polosukhin, ``Attention is all you need,''
  \emph{Advances in neural information processing systems}, vol.~30, 2017.

\bibitem{hendrycks2016gaussian}
D.~Hendrycks and K.~Gimpel, ``Gaussian error linear units (gelus),''
  \emph{arXiv preprint arXiv:1606.08415}, 2016.

\bibitem{han2023agent}
D.~Han, T.~Ye, Y.~Han, Z.~Xia, S.~Song, and G.~Huang, ``Agent attention: On the
  integration of softmax and linear attention,'' \emph{arXiv preprint
  arXiv:2312.08874}, 2023.

\bibitem{wang2004image}
Z.~Wang, A.~C. Bovik, H.~R. Sheikh, and E.~P. Simoncelli, ``Image quality
  assessment: from error visibility to structural similarity,'' \emph{IEEE
  transactions on image processing}, vol.~13, no.~4, pp. 600--612, 2004.

\bibitem{mayer2016large}
N.~Mayer, E.~Ilg, P.~Hausser, P.~Fischer, D.~Cremers, A.~Dosovitskiy, and
  T.~Brox, ``A large dataset to train convolutional networks for disparity,
  optical flow, and scene flow estimation,'' in \emph{Proceedings of the IEEE
  conference on computer vision and pattern recognition}, 2016, pp. 4040--4048.

\bibitem{geiger2012we}
A.~Geiger, P.~Lenz, and R.~Urtasun, ``Are we ready for autonomous driving? the
  kitti vision benchmark suite,'' in \emph{2012 IEEE Conference on Computer
  Vision and Pattern Recognition}.\hskip 1em plus 0.5em minus 0.4em\relax IEEE,
  2012, pp. 3354--3361.

\bibitem{menze2015object}
M.~Menze and A.~Geiger, ``Object scene flow for autonomous vehicles,'' in
  \emph{Proceedings of the IEEE conference on computer vision and pattern
  recognition}, 2015, pp. 3061--3070.

\bibitem{scharstein2014high}
D.~Scharstein, H.~Hirschm{\"u}ller, Y.~Kitajima, G.~Krathwohl,
  N.~Ne{\v{s}}i{\'c}, X.~Wang, and P.~Westling, ``High-resolution stereo
  datasets with subpixel-accurate ground truth,'' in \emph{Pattern Recognition:
  36th German Conference, GCPR 2014, M{\"u}nster, Germany, September 2-5, 2014,
  Proceedings 36}.\hskip 1em plus 0.5em minus 0.4em\relax Springer, 2014, pp.
  31--42.

\bibitem{schops2017multi}
T.~Schops, J.~L. Schonberger, S.~Galliani, T.~Sattler, K.~Schindler,
  M.~Pollefeys, and A.~Geiger, ``A multi-view stereo benchmark with
  high-resolution images and multi-camera videos,'' in \emph{Proceedings of the
  IEEE conference on computer vision and pattern recognition}, 2017, pp.
  3260--3269.

\bibitem{adamw}
I.~Loshchilov and F.~Hutter, ``Fixing weight decay regularization in adam,''
  2018.

\bibitem{stommel2013inpainting}
M.~Stommel, M.~Beetz, and W.~Xu, ``Inpainting of missing values in the kinect
  sensor's depth maps based on background estimates,'' \emph{IEEE Sensors
  Journal}, vol.~14, no.~4, pp. 1107--1116, 2013.

\bibitem{menze2015joint}
M.~Menze, C.~Heipke, and A.~Geiger, ``Joint 3d estimation of vehicles and scene
  flow,'' \emph{ISPRS annals of the photogrammetry, remote sensing and spatial
  information sciences}, vol.~2, pp. 427--434, 2015.

\bibitem{zheng2023diffuvolume}
D.~Zheng, X.-M. Wu, Z.~Liu, J.~Meng, and W.-s. Zheng, ``Diffuvolume: Diffusion
  model for volume based stereo matching,'' \emph{arXiv preprint
  arXiv:2308.15989}, 2023.

\bibitem{song2020edgestereo}
X.~Song, X.~Zhao, L.~Fang, H.~Hu, and Y.~Yu, ``Edgestereo: An effective
  multi-task learning network for stereo matching and edge detection,''
  \emph{International Journal of Computer Vision}, vol. 128, no.~4, pp.
  910--930, 2020.

\bibitem{shen2021cfnet}
Z.~Shen, Y.~Dai, and Z.~Rao, ``Cfnet: Cascade and fused cost volume for robust
  stereo matching,'' in \emph{Proceedings of the IEEE/CVF Conference on
  Computer Vision and Pattern Recognition}, 2021, pp. 13\,906--13\,915.

\end{thebibliography}
\bibliographystyle{IEEEtran}

\vfill

\end{document}